\PassOptionsToPackage{dvipsnames,table}{xcolor}
\documentclass{article} %
\usepackage{iclr2027_conference,times}
\iclrfinaltrue

\usepackage{amsmath,amsfonts,bm}

\def\eqref#1{equation~\ref{#1}}

\def\1{\bm{1}}

\DeclareMathAlphabet{\mathsfit}{\encodingdefault}{\sfdefault}{m}{sl}
\SetMathAlphabet{\mathsfit}{bold}{\encodingdefault}{\sfdefault}{bx}{n}

\usepackage{float} 

\usepackage{wrapfig}

\usepackage{graphicx} %
\usepackage[table]{xcolor}
\usepackage{colortbl}

\usepackage{tikz}
\usetikzlibrary{arrows.meta,positioning}

\usepackage{amsmath, amssymb, amsthm}
\usepackage{upgreek}

\usepackage{booktabs}
\usepackage{longtable}

\usepackage{algorithm}
\usepackage[noend]{algpseudocode}

\usepackage{hyperref}
\usepackage{url}

\newtheoremstyle{boldnote}
  {6pt}{-3pt}
  {\itshape}{}
  {\bfseries}{.}
  {5pt plus 1pt minus 1pt}
  {\thmname{#1}\thmnumber{ #2}\thmnote{ \bfseries(#3)}}
\theoremstyle{boldnote}

\newtheorem{definition}{Definition} 
\newtheorem{remark}{Remark}
\newtheorem{assumption}{Assumption}

\newtheoremstyle{boldnoteup}
  {6pt}{-3pt}
  {\normalfont}{}
  {\bfseries}{.}
  {5pt plus 1pt minus 1pt}
  {\thmname{#1}\thmnumber{ #2}\thmnote{ \bfseries(#3)}}
\theoremstyle{boldnoteup}
\newtheorem{example}{Example}

\providecommand{\Pa}{\operatorname{Pa}}

\providecommand{\cdag}{\text{C-DAG}}
\providecommand{\acdag}{\ensuremath{\alpha\text{C-DAG}}}
\providecommand{\accpdag}{\ensuremath{\alpha\text{C-CPDAG}}}
\providecommand{\olaca}{GO4CD}
\providecommand{\cloc}{CLOC}

\newcommand{\ind}{\perp\!\!\!\perp}

\title{GO-Based Clustering for Learning Cluster-Level Causal Gene Regulatory Networks}

\author{Azlaan Mustafa Samad\\
L3S Research Center, Leibniz University Hannover, \\ 
Lower Saxony Center for Artificial Intelligence and Causal Methods in Medicine (CAIMed),\\
Hannover, Germany \\
\texttt{azlaan.samad@l3s.de} \\
\And
Wei Zhang \\
L3S Research Center, \\
Leibniz University Hannover, Germany \\
\texttt{wei.zhang@l3s.de} \\
\AND
Adèle H Ribeiro \\
Institute for Causal and Explainable Artificial Intelligence in Life Sciences, \\
Center for Computational Life Sciences,\\
RWTH Aachen University, Germany \\
\texttt{adele.ribeiro@rwth-aachen.de}
}

\begin{document}
\raggedbottom

\maketitle

\begin{abstract}
Discovery of causal relationships in high-dimensional Gene Regulatory Networks (GRN) is computationally challenging and often difficult to interpret due to dense connections. Therefore, grouping genes together into functional modules can improve tractability and biological interpretability. However, existing cluster level causal discovery methods assume access to a predefined \emph{admissible} partition, requiring the graph over clusters to be acyclic. 
Constructing such partitions is therefore challenging.
In this work, we introduce GO-based Clustering for Causal Discovery (\textbf{GO4CD}), an algorithm that uses Gene Ontology (GO)
to construct biologically meaningful gene partitions at multiple levels of granularity, while favoring those more likely to be admissible for causal discovery. 
GO4CD  groups together genes participating in a shared biological process, and propagates gene annotations through the ontology hierarchy to achieve different granularity of partitions. 
Furthermore, we integrate GO4CD with Causal Learning over Clusters (CLOC) algorithm and evaluate recovery of true Markov equivalence class both with an oracle of conditional independencies and on simulated gene expression data using multivariate conditional independence tests. We evaluate GO4CD on multiple \emph{E.coli} regulatory subnetworks and find that it is \emph{inadmissible} in 18.1\% of the cases, 
compared with 65.3-82.3\% for the semantic-similarity baselines. Our results indicate that GO4CD is substantially better suited to learning causal GRNs defined over biologically meaningful gene clusters.
\end{abstract}

\section{Introduction}

Causal discovery aims to discover cause-and-effect relationship between variables and therefore finds application across multiple domains~\citep{pearl2003causality}. Understanding the physical and chemical interactions within a cell is a critical step for developing targeted therapies for complex diseases~\citep{tejada2025causal}. These interactions are typically modeled using Gene Regulatory Networks (GRNs), which map genes as nodes and their regulatory relationships as directed edges. While traditional methods frequently infer GRNs by fitting models to gene co-expression, these approaches increasingly produce correlational "hairballs" that fail to capture true molecular causation~\citep{hairball}. Inferring causal GRNs where edges strictly represent mechanistic cause-and-effect relationships has become a central challenge in computational genomics.

While gene knockout experiments are capable of uncovering direct causal relationships, inferring a complete causal network at the individual gene level is practically infeasible due to the sheer volume of combinatorial interventions required~\citep{scheines2001constructing}. Furthermore, transcriptomics datasets suffer from the curse of dimensionality, where the number of measured genes outnumbers the available samples~\citep{curseofdimensionailtyGRN}. Consequently, the dimensionality of the structural search space is astronomically large; even a small network of just ten genes possesses over $10^{30}$ possible network topologies~\citep{hairball}. 
The large number of genes (approx. 20,000 in humans)~\citep{international2004finishing} and the dense connections between them make recovering exact causal connections in GRNs %
statistically
and computationally intractable, as well as difficult to interpret. 
Intuitively, however, biological systems do not operate as flat networks of isolated genes. Rather, they are organized into functional modules and hierarchies that bridge the gap from molecular components to cellular phenotypes~\citep{hairball}. %
To address this, biologists coarse-grain the network to a ``representational'' level by grouping co-expressed genes into functional modules~\citep{segal2003module}. 
Once genes are clustered, one can aim to identify a coarser causal diagram capturing relationships among broader biological mechanisms.

Grouping 
variables
into clusters is challenging because the resulting partitions must make biological sense while simultaneously satisfying the strict %
assumptions of causal discovery. A major bottleneck here is the concept of \emph{admissibility}, which requires the resulting 
graph  between clusters to be entirely \emph{acyclic}. This does not naturally follow from the gene network being acyclic, since two clusters can each contain an ancestor of a gene in the other, thus closing a cycle that never existed between individual genes. To find these admissible partitions, state-of-the-art methods depend on fully specified Directed Acyclic Graphs (DAGs) over all variables~\citep{r20, r21}. Others that learn from data do so under overly restrictive assumptions such as interventional data~\citep{chalupka2016unsupervised}, linear functional forms~\citep{r19} or the absence of latent confounders~\citep{ r15}.
Other methods~\citep{ruizdevargas, cloc} simply assume the clusters are already provided.
Satisfying these requirements from data alone is fundamentally difficult across domains, but it is especially a problem in biology.

To tackle this open problem, we introduce \textbf{GO-based Clustering for Causal Discovery (\olaca{})}, an algorithm that leverages prior knowledge from the Gene Ontology (GO) to partition the genes into clusters with biological meaning which are more likely to be admissible for causal discovery. Using the ontology hierarchy, \olaca{} can construct these partitions at different levels of granularity.
\olaca{} achieves this by grouping genes based on shared biological processes and systematically propagates these annotations through the ontology hierarchy.
\olaca{} explicitly injects established biological priors to guide the abstraction process, making it far less likely to group genes in a way that creates spurious cluster-level cycles.
We then integrate \olaca{} with the Causal Learning over Clusters (CLOC)~\citep{cloc} algorithm to evaluate \cloc{}'s performance under two settings: oracle mode which guarantees perfect Conditional Independence (CI) assessment and on the gene expression data using multivariate CI tests. We test \cloc{} under oracle mode to report its performance solely based on inadmissible partitions by eliminating the effect of imperfect CI test, since under admissible partition and perfect independence test, \cloc{} guarantees recovery of the \emph{Markov equivalence class} (MEC) as it is sound and complete. Furthermore, we also test \cloc{} on simulated gene expression data using multivariate CI tests where the errors are compounded from imperfect CI tests and inadmissible partitions for practical application.
Extensive evaluations on \emph{E. coli} regulatory subnetworks demonstrate that
\olaca{}'s partitions are inadmissible in only $18.1\%$ of the cases, against
$65.3\%$ to $82.3\%$ for standard semantic-similarity baselines. To our knowledge, no existing method reports whether the partitions it uses are admissible. Ultimately, our results show that \olaca{} is much better equipped to learn causal GRNs across biologically meaningful gene groups.

\section{Related Work}

\paragraph{Causal Abstraction and Discovery over Variable Clusters}

Causal abstraction asks when low- and high-level causal models are consistent under corresponding interventions~\citep{r24,r6,r7}. 
Work on clusters either takes the partition as given or constructs it.
\cite{r8} and \cite{cloc} require a supplied \emph{admissible} partition, inducing no directed cycles between clusters, while \cite{r16} assume predefined groups and an acyclic group-level model.
\cite{r17} optimize aggregation within predefined groups.
\cite{ruizdevargas} use a supplied C-DAG, edges included, as background knowledge for variable-level causal discovery.
Where partitions or abstraction mappings are learned, the methods differ in their required inputs.
\cite{r15} and \cite{rajpal} build clusters from observational data.
\cite{r19} learn linear abstraction mappings to specified high-level variables using paired low and high-level observations.
\cite{r18} learn abstract neural causal models using supplied structural knowledge.
\cite{zennaro} learn domain mappings between supplied causal models using interventional distributions.
\cite{r20} and \cite{r21} instead cluster a known causal DAG, the former preserving causal-effect identifiability.
In closely related work, \citet{r23} construct partitions and acyclic cluster-level graphs from observational and interventional data with unknown intervention targets.
\olaca{} builds gene partitions from GO knowledge without using expression data or requiring a known variable-level causal graph.

\paragraph{Gene Ontology and Semantic-Similarity-Based Clustering}

Semantic-similarity-based clustering is widely used in natural language processing~\citep{r25,r26} and has also been applied in biology.
GO~\citep{go2000} provides a structured vocabulary and gene-product annotations for biological processes, molecular functions, and cellular components. 
Semantic-similarity measures are developed to quantify functional relatedness.
\cite{resnik1995using} relies on the most informative common ancestor, while \cite{linmethod} normalize the shared information from the compared terms. 
\cite{wangmethod} use ancestor contributions and GO relation types to derive similarities between GO terms and genes.
A separate line of work represents GO terms and genes as embeddings, derived either strictly from ontology structure~\citep{anc2vec} or jointly with gene annotations~\citep{onto2vec}. We do not compare against these directly since the two sources they learn from are the same ones our baselines use in closed form, the ontology's structure in \cite{wangmethod} and annotation frequencies in \cite{resnik1995using} and \cite{linmethod}. Introducing embeddings would merely test a parameterized approximation of the same fundamental ontology signals already represented in our study.

\section{Preliminaries}\label{sec:prelim}
In this section, we first review the basic concepts of causal directed acyclic graphs (DAGs), then their abstracted form \emph{cluster-DAG} (\cdag{}) and finally the \emph{Causal Learning Over Clusters} (\cloc{}) algorithm.  

We assume the underlying causal structure is a standard DAG, $G(\mathbf{V}, E)$, over a set of variables $\mathbf{V}$. In a nutshell, constraint based causal discovery  algorithms look for marginal/conditional independencies to prune and orient the graph. These (in)dependencies are encoded through \emph{d-separation}. We operate under \emph{faithfulness} assumption, which states that any independencies in the data imply separation in the graph. However, under observational data setting, certain graphs cannot be distinguished since they carry the same independencies, for e.g. $X \to Z \to Y$, $X \leftarrow Z \to Y$ and
$X \leftarrow Z \leftarrow Y$ all say only that $X$ and $Y$ are independent given $Z$. DAGs that share a skeleton and the same v-structures
form a MEC, and observational data identifies the causal
structure only up to that class. The class is drawn as a Completed Partially Directed Acyclic Graph (CPDAG), which directs the edges every member agrees on and leaves the rest undirected.
To formalize the structural abstraction of a DAG, we build upon the framework introduced by \cite{cloc}. We first define how a DAG translates into a \cdag{} below.
\begin{definition}[Cluster DAG (C-DAG) \citep{cloc}]
\label{def:cdag}
Let $G(\mathbf{V}, E)$ be a DAG over a vertex set $\mathbf{V}$, and let
$\mathcal{C} = \{C_1, \dots, C_k\}$ be a partition of $\mathbf{V}$. Construct a
graph $G_{\mathcal{C}}(\mathcal{C}, E_{\mathcal{C}})$ over $\mathcal{C}$ with edge
set $E_{\mathcal{C}}$ defined as follows: an edge $C_i \rightarrow C_j$ is in
$E_{\mathcal{C}}$ if and only if there exist $V_i \in C_i$ and $V_j \in C_j$
such that $V_i \in \Pa(V_j)$ in $G$. If $G_{\mathcal{C}}(\mathcal{C}, E_{\mathcal{C}})$ contains no directed cycles, we say that $\mathcal{C}$ is an
\textbf{admissible partition} of $\mathbf{V}$, and we call $G_{\mathcal{C}}$ a
\emph{cluster DAG}, or \cdag{}, compatible with $G$.
\end{definition}

\begin{remark}\label{rem:trivial_acyclic}
    Obtaining a non-trivial \textit{admissible} partition is not straightforward. Taking every $v \in \mathbf{V}$ as its own cluster is admissible but abstracts
    nothing, while collecting all the variables into a single cluster yields a
    graph with one node and no edges. Furthermore, acyclicity of $G$ does not carry over to $G_{\mathcal{C}}$. For e.g. two clusters can each contain a parent of a variable in the other, say $V_i \to V_j$ with $V_i \in C_1, V_j \in C_2$ and $V_j' \to V_i'$ with $V_j' \in C_2, V_i' \in C_1$, which closes a cycle $C_1 \leftrightarrows C_2$ in $G_{\mathcal{C}}$ while $G$ stays acyclic. Then such a construction does not yield \cdag{} at all, since $\mathcal{C}$ is not admissible. 
\end{remark}

\cite{cloc} also introduced a new graphical representation to accurately represent the (in)dependencies over a \cdag{}, formally known as \acdag. Similar to DAG, \acdag{} are recoverable up to a MEC known as \accpdag{}. Under \emph{faithfulness} assumption, one can recover the \accpdag{} using the \emph{Causal Learning over Clusters} (CLOC) algorithm, given an \emph{admissible} partition~(Algorithm 1 in \cite{cloc}). The \cloc{} algorithm operates in three phases: separating, marking and the orientation phase. In the separation phase, it removes edges for a pair of nodes $\mathbf{X}$ and $\mathbf{Y}$ if it finds a separation set $\mathbf{S}$ such that ($\mathbf{X}\ind\mathbf{Y}\mid\mathbf{S}$). In the second phase, marks are added over the triplets to encode (in)dependencies and finally the orientation phase which orients edges.

\begin{remark}
\label{rem:admissibility}
Whether a partition is admissible depends on $G$, the DAG over $\mathbf{V}$, which is unknown. To
check it one would need the edges of $G$, and learning those is what clustering is meant to avoid.
CLOC neither tests the condition nor constructs a partition that satisfies it, rather it assumes an admissible partition to be handed over.
\end{remark}

This constitutes a significant bottleneck in applying \cloc{} to real-world scenarios. In Section~\ref{sec:olaca}, we introduce our core contribution, a methodology to construct admissible partitions using the background knowledge described in Section~\ref{sec:GO}, which we then apply to the \cloc{} algorithm.

\section{Background Knowledge}\label{sec:GO}
Remark~\ref{rem:trivial_acyclic} and~\ref{rem:admissibility} open the problem of forming non-trivial admissible partition. Given a set of genes, there are many ways to partition them, from every single gene as its own cluster, to all of them collapsed to a single blob. We use Gene Ontology to restrict the space of clusters, which we describe in this section, as a necessary pre-requisite for our clustering algorithm.

\textbf{Gene Ontology (GO).} GO is a standardised, structured and species-agnostic representation of biological knowledge \citep{go2000,GO}. It is
organised in three parts: \emph{Molecular Function (MF), Cellular Component (CC) and
Biological Process (BP).} As the name suggests, MF are activities performed by
either a single or multiple gene products, and CC refer to the locations where
these activities are performed. BPs are large, complex biological (or gene)
programs made up of multiple MFs. For example, the MF term
\emph{cyclin-dependent protein kinase activity} is part of the BP
\emph{regulation of cell cycle}. We use only BP in this work, because
it describes gene programs, and from here on we use the words \emph{term} and \emph{biological process} interchangeably.

\textbf{Relations in GO.} GO arranges these terms as a DAG, where the terms are the
nodes and the relations between them are the edges. The backbone is made of
\texttt{is\_a} and \texttt{part\_of} relations: A \texttt{is\_a} B means A is a type of B, and A \texttt{part\_of} B means A is a sub-process of B, so A never occurs outside B. The parent terms are broader than
the child terms, which are more specialised. Unlike a strict hierarchy, where a
child has only a single parent, GO allows a child to have more than one parent.
For example, \emph{autophagosome assembly} has two parents, \emph{organelle
assembly} and \emph{autophagosome organization}. \emph{Biological Process} is the
root of this branch, and all the other BP terms like \emph{biological regulation}
and \emph{cellular process} sit below it. On top of this backbone GO also records
\texttt{regulates} relations, which connect one process to another it controls.
These do not follow the hierarchy, and they are the source of the edges between
our clusters.

\textbf{Granularity in GO.} As one goes down a branch the terms become more specific, referring to a specialised type of BP. Since GO is a DAG, a term can have multiple parents and thus multiple paths to the root node. On top of this, not all parents sit at the same distance from the root node. Thus different paths to the root node can have different lengths. We use length of the longest such paths as its \textbf{depth}. For \textit{autophagosome assembly}'s two parents, \textit{organelle assembly} sits at depth 5 and \textit{autophagosome organization} at depth 6, so its depth is 6. 
In GO, some BPs are far more studied than others and have therefore been broken down into many child terms. For example, \emph{regulation of metabolic process} and \emph{ferroptosis} both sit at depth 4, but the former has 2,405 terms below it while the latter is a leaf. This means that two terms sitting at the same depth can still describe processes of very different breadth. We nonetheless use depth as our measure of specificity, and we state this as an assumption rather than as a property of GO.

\begin{assumption}[Depth as specificity]
\label{asm:depth-specificity}
For any two terms $T$ and $T'$ with $\mathrm{depth}(T) = \mathrm{depth}(T')$, $T$ and $T'$ are equally specific.
\end{assumption}

\textbf{Gene annotations in GO.} A biological process is carried out by a set of genes acting together, and genes/gene products are connected to GO terms via annotations. A gene is annotated to BPs mainly with two labels: \texttt{involved\_in}, if it plays an integral role in that process, that is, the gene is a \textbf{participant} of the term; and \texttt{acts\_upstream\_of}, if its activity occurs upstream of the process, that is, the gene is a \textbf{regulator} of the term.

In summary, GO terms represent distinct biological processes, their hierarchical depth indicates their level of specificity, regulation relations encode the interactions between these processes, and annotations map individual genes to the processes in which they participate.

\section{Go4CD: GO-based Clustering for Causal Discovery}\label{sec:olaca}
Fixing a depth fixes the granularity, but it does not hand us a partition. At a given depth, a gene can take part in several terms, so their participant sets  overlap.
We therefore group two genes together whenever some term at that depth has both of them as participants, and we take the clusters to be the connected components of this grouping, so no term's participants end up in two different clusters. 

\textbf{Notation.} Fix an ontology depth $d$. We write $\mathcal{T}_d$ for the
terms at depth $d$, $\mathcal{T}_{\geq d}$ for the terms at depth $d$ or deeper,
and $\mathrm{part}(T)$ for the \emph{participant} genes of
$T \in \mathcal{T}_{\geq d}$. The genes \emph{active} (or annotated) at depth $d$
are $V_d = \bigcup_{T \in \mathcal{T}_d} \mathrm{part}(T)$, and a gene outside
$V_d$ is one whose annotations are all shallower than $d$, we leave it as a singleton cluster. We
write $\Pi(\mathbf{V})$ for the partitions of the gene set $\mathbf{V}$, where a partition
$\{C_1,\dots,C_k\}$ has non-empty, pairwise-disjoint
\textbf{clusters} covering $\mathbf{V}$, and
$\mathcal{R}_d \subseteq \mathcal{T}_{\geq d} \times \mathcal{T}_{\geq d}$ for the regulation edges, the ordered pairs $(s,t)$ with $s$ \texttt{regulates} $t$, where $s$ and $t$ are terms in GO.

\begin{example}\label{ex:coparticipation} Let the terms at depth $d$ be
$T_1, T_2, T_3$ with $\mathrm{part}(T_1) = \{g_1,g_2\}$,
$\mathrm{part}(T_2) = \{g_2,g_3\}$ and $\mathrm{part}(T_3) = \{g_4,g_5\}$.
Co-participation gives $C_1 = \{g_1,g_2,g_3\}$ and $C_2 = \{g_4,g_5\}$, and
splitting $C_1$ to isolate $g_3$ would tear $T_2$.
\end{example}

\begin{definition}[Term Integral Partition]
    \label{def:term-integral}
A partition $\mathcal{C} \in \Pi(\mathbf{V})$ is \textbf{term-integral} if every
term's participant set lies inside a single cluster:
\begin{equation*}
  \forall\, T \in \mathcal{T}_d \;\; \exists\, C \in \mathcal{C} \;:\;
  \mathrm{part}(T) \subseteq C .
\end{equation*}
\end{definition}

That is what makes a cluster worth reading, if the participants of a term ended up in two different clusters, neither would stand for that process, and an edge into either of them could not be read as an edge into the process. Thus far, we described a single depth where \textit{every} gene is annotated to a term at that depth. However in reality, most genes are not, since a gene is annotated to the most specific term the evidence supports. The \textit{True Path Rule} (Definition~\ref{TPR}) is what lifts them up, because a gene annotated to a term is also a \textbf{participant} of every broader (parent) process above it. A gene annotated at depth 9 therefore counts as a participant of its depth 4 ancestors as well, and this is what leaves genes attached to the terms at whichever depth we cut. The same lifting applies to \textbf{regulators}.

\begin{definition}[True Path Rule]\label{TPR}
If a gene $g$ is a participant of a term $T'$, then $g$ is a participant of every
ancestor $T$ of $T'$ along \texttt{is\_a} and \texttt{part\_of}:
\begin{equation*}
  \bigcup_{T' \,\prec\, T} \mathrm{part}(T') \;\subseteq\; \mathrm{part}(T),
  \qquad \text{where } T' \prec T \text{ means } T' \text{ is a descendant of } T .
\end{equation*}
\end{definition}

From here on $\mathrm{part}(T)$ is understood to include the genes lifted from $T$'s descendants in this way, so both $V_d$ and the clusters are computed from these enlarged participant sets.

To carry the regulation relations over from terms to clusters we use $\kappa_{\mathcal{C}}(T)$ to map from a term to its cluster. By the True Path Rule the participants of a deeper term lie inside those of its depth-$d$ ancestors, so a term-integral partition places every term of $\mathcal{T}_{\geq d}$ in a single cluster. A regulation edge survives the coarsening when its two terms land in different clusters, and it is absorbed when they land in the same one. We call these \textbf{GO regulation edges}, since they come from the ontology itself. Another source of edges is when a gene is a regulator of a term while participating in some other term, we call these \textbf{Annotation edges}.

\begin{example}\label{ex:regedge}
Continuing Example~\ref{ex:coparticipation}, let
$\mathcal{R}_d = \{(T_1,T_2), (T_1,T_3), (T_2,T_3)\}$. Here
$\kappa_{\mathcal{C}}(T_1) = \kappa_{\mathcal{C}}(T_2) = C_1$ and
$\kappa_{\mathcal{C}}(T_3) = C_2$, so $(T_1,T_2)$ is absorbed inside $C_1$ while
the other two survive as the single edge $C_1 \rightarrow C_2$. If $g_4$ is also
a regulator of $T_1$, the annotation edge $C_2 \rightarrow C_1$ appears and the
two clusters point at each other.
\end{example}

\begin{definition}[Cluster GO Graph]
\label{def:cluster-graph}
Let $\mathcal{C}$ be term-integral, $\mathcal{R}_d$ the set of GO term pairs $(s,t)$ with \texttt{regulates} edge $s \rightarrow t$, and $\kappa_{\mathcal{C}} : \mathcal{T}_{\geq d} \rightarrow \mathcal{C}$
a mapping from a term $T$ to its cluster in $\mathcal{C}$. Construct a
graph $H_{\mathcal{C}}(\mathcal{C}, E_{H})$ over $\mathcal{C}$ with edge set $E_{H}$ defined as
follows: an edge $C \rightarrow C'$, $C \neq C'$, is in $E_{H}$ if and only if some
$(s,t) \in \mathcal{R}_d$ has $\kappa_{\mathcal{C}}(s) = C$ and $\kappa_{\mathcal{C}}(t) = C'$
(a \emph{GO regulation edge}), or some gene $g \in C$ is annotated
\texttt{acts\_upstream\_of} a term $T$ with $\kappa_{\mathcal{C}}(T) = C'$ (an
\emph{annotation edge}).
\end{definition}

Coarsening can create \emph{cycles} in $H_{\mathcal{C}}$ for the same reason it does in a coarsened graph (Section~\ref{sec:prelim}), one term in a cluster regulates a term in another while a different term or gene there acts back (see Example~\ref{ex:regedge}).

A \cdag{} exists only when $\mathcal{C}$ is \emph{admissible}, that is, when the coarsened gene-level DAG $G_{\mathcal{C}}$ is acyclic (Definition~\ref{def:cdag}).
To remove the cycles, we contract the \textbf{Strongly Connected Components} (SCC) of $H_{\mathcal{C}}$ into single clusters, which merges the clusters on the cycle and keeps the partition term-integral (Definition~\ref{def:term-integral}), since a merge is a union of clusters. Thus, in Example~\ref{ex:regedge} merging will leave a single cluster $\{g_1,\dots,g_5\}$.

We call this term-integral, depth aware construction \textbf{GO for Causal Discovery (\olaca{})} summarised in Algorithm~\ref{alg:olaca}. 

\vspace{-2mm}
\begin{algorithm}[H]
\setlength{\abovecaptionskip}{0pt}
\caption{\olaca{}: Term-integral partition at depth $d$}
\label{alg:olaca}
\begin{algorithmic}[1]
\Require GO DAG $\mathcal{O}$, annotation table $A$, gene set $\mathbf{V}$, depth $d$
\State participants and regulators of each $T \in \mathcal{T}_d$ from $A$, then lifted from every deeper term \Comment{Def.~\ref{TPR}}
\State $\mathcal{C} \gets$ connected components of co-participation on $V_d = \bigcup_{T \in \mathcal{T}_d} \mathrm{part}(T)$, one singleton per gene of $\mathbf{V} \setminus V_d$
\State $\kappa_{\mathcal{C}}(T) \gets$ the cluster holding $\mathrm{part}(T)$, and add $C \rightarrow C'$ to $H_{\mathcal{C}}$ for every GO regulation edge or annotation edge running from $C$ to $C'$ \Comment{Def.~\ref{def:cluster-graph}}
\While{$H_{\mathcal{C}}$ has a directed cycle}
  \State contract each SCC; rebuild $\kappa_{\mathcal{C}}, E_{H}, H_{\mathcal{C}}$
\EndWhile
\State \Return $\mathcal{C}, H_{\mathcal{C}}$
\end{algorithmic}
\end{algorithm}
\vspace{-4mm}

\begin{remark}
    \noindent If GO records all edges and annotations at every depth, every edge of the \cdag{} is also an edge of $H_{\mathcal{C}}$, so merging the SCC of $H_{\mathcal{C}}$ leaves both graphs acyclic. But GO does not cover everything, so \cdag{} can have edges that $H_{\mathcal{C}}$ does not.
\end{remark}

\begin{example}\label{ex:cdag-cycle}
Let $C_1 = \{g_1, g_2\}$ and $C_2 = \{g_3, g_4\}$, and let the gene network contain $g_1 \rightarrow g_3$ and $g_4 \rightarrow g_2$. No gene lies on a cycle, but the C-DAG has both $C_1 \rightarrow C_2$ and $C_2 \rightarrow C_1$. If GO records neither relation, $H_{\mathcal{C}}$ carries no edge between the two clusters at all, so the merge sees an acyclic graph and leaves the cycle standing.
\end{example}

This is a limitation of GO's knowledge, merge can only remove the cycles GO knows about. How often this leads to violation of admissibility, we measure in Section~\ref{sec:admissibility}.

\section{Experiments}\label{sec:experiments}
We investigate two core questions: how often \olaca{} produces admissible partitions, and whether \cloc{} achieves better structural recovery on these partitions compared to semantic baselines. We evaluate \cloc{} in two settings. In \emph{oracle} mode, we derive (in)dependencies directly via d-separation on the underlying gene-level DAG. In \emph{data} mode, dependencies are inferred through conditional independence (CI) tests on expression data, which introduces statistical noise. To support this second mode, we extend the standard variable-level \cloc{} implementation to a multivariate setting.
\vspace{-1mm}

\subsection{Setup}

\textbf{Networks and Sampling.}\label{sec5:setup} Our evaluation relies on a
curated \emph{E.\ coli} gene regulatory network from
RegulonDB~\citep{salgado2023regulondb}. We strip out all feedback edges, so
the gene-level graph is acyclic and any cluster-level cycle we report comes
from the coarsening rather than from a loop we inherited. From this network
we extract nested random subnetworks at seven sizes ($50$ to $1000$ genes)
with five seeds per size, giving $35$ subnetworks, each grown outward from a
random seed gene following \citet{marbach2009generating}
(Appendix~\ref{app:samplesubnet}). RegulonDB holds only $1781$ genes, so the
replicates overlap as they grow, sharing $2$ genes at size $50$ and $839$ at
size $1000$ (details in Appendix~\ref{app:samplesubnet}).

\textbf{Baselines.} We compare against three semantic-similarity measures
that read the same GO annotations as \olaca{} but parse them differently.
\textbf{Resnik}~\citep{resnik1995using} scores a gene pair by the
information content of their most informative common ancestor,
\textbf{Lin}~\citep{linmethod} normalises that score, and
\textbf{Wang}~\citep{wangmethod} ignores annotation counts and instead
propagates a discounted score through a term's ancestors. All three return a
gene-by-gene similarity matrix rather than a partition, so we cluster each
one with average linkage and cut the dendrogram at the $k$ that \olaca{}
produces at that depth (details in Appendix~\ref{app:baseline}) .

\textbf{Comparison Criteria.} We cluster each subnetwork across every depth of the ontology. To ensure comparisons are made at identical granularities, the baseline methods are constrained to match the exact cluster count generated by \olaca{} at the corresponding depth. Comparisons are conducted strictly instance-wise, where an \emph{instance} is defined by a fixed subnetwork size ($n$), replicate, and ontology depth. Consequently, all four methods operate on the exact same set of genes and output the same number of clusters ($k$). In total, this yields 288 distinct instances (e.g., an $n=50$ subnetwork evaluated at depth 7 and partitioned into $k=20$ clusters represents one instance). We discard instances where $k = n$, as each gene trivially forms its own cluster, collapsing the \cdag{} back to the inherently acyclic gene-level DAG and yielding identical results across all methods.

\textbf{Dataset Generation.} We utilized CausalRegNet~\citep{kovavcevic2024simulation} to generate realistic gene expression data for all subnetworks based on their underlying \emph{acyclic} GRNs. We selected this model because it scales efficiently, copes with the issue of varsortability, and produces data that is both \emph{faithful} to the underlying graph and consistent with the key distributional properties of real-world data. Full generation details are available in Appendix~\ref{app:datagen}.

\textbf{CI test.} Tests at the cluster level are multivariate, since a cluster carries as many
dimensions as it has genes. Kernel tests are too slow at this scale, so we use the Generalised
Covariance Measure \citep{gcmcovarincemeasure} instead, which regresses $\mathbf{X}$ and
$\mathbf{Y}$ on $\mathbf{Z}$ and asks whether the residuals stay uncorrelated. It applies at any dimension and needs no kernel matrix. We use its multivariate form with a Gaussian multiplier
bootstrap, and take residuals from random forests with \texttt{min\_samples\_leaf} set to 10, and test at a significance level of $0.05$.

\textbf{Implementation Details.} Due to computational constraint, we run \cloc{} only on instances with at most 60 clusters in the \emph{oracle} mode, and in \emph{data} mode only on the three smallest network sizes ($50$, $100$ and $200$ genes) with at most 20 clusters and separating sets of size at most 2.

\subsection{Admissibility of the Constructed Partitions}
\label{sec:admissibility}
\olaca{} relies entirely on GO as prior knowledge, enabling its application to networks with unknown regulatory edges. However, because GO does not annotate every gene at every depth, the extent to which this knowledge guides \olaca{} toward an \emph{admissible} partition remains an empirical question, which we evaluate here.

\textbf{\olaca{} leads with more admissible partitions than any baseline across all subnetworks.}
In Table~\ref{tab:premerge-admissibility}, we report the percentage of instances whose \cdag{} is cyclic. Each row pools the five replicates and the depths where all the methods return the same $k$, and the last row pools every size. Overall, \olaca{} is inadmissible on just 18.1\% of instances while all other baselines range between $65.3\%$--$82.3\%$. \olaca{} also stays the lowest at every size. Even at smaller sizes of $50$ and $100$, where the five replicates share only $2$ and $47$ genes respectively, \olaca{} is inadmissible on $10.0\%$ and $13.2\%$ of instances against $32.5$--$52.6\%$ for the baselines.

\textbf{The GO-informed gene distribution explains the admissibility.} To test whether \olaca{}'s  partition admissibility stems from its cluster size distribution or its GO-informed gene placement, we introduce a \emph{Rand} baseline that randomly assigns genes while strictly preserving \olaca{}'s exact cluster sizes (averaged over ten draws). While \emph{Rand} perfectly mimics \olaca{}'s structural shape, it remains cyclic $80.9\%$ of the time, producing only $3$ completely acyclic instances out of $288$. Performance-wise, Resnik \textbf{(R)} aligns with \emph{Rand}, Lin \textbf{(L)} and Wang \textbf{(W)} exceed it by $15$ and $12$ percentage points respectively, and \olaca{} outperforms it by $63$ points. This confirms that admissibility is driven by GO's gene grouping rather than partition topology (see Table~\ref{tab:kbin-shape} in Appendix~\ref{app:olaca_admiss}).

As shown in Table~\ref{tab:premerge-admissibility}, \olaca{} typically concentrates genes into one dominant cluster alongside many singletons, whereas baselines produce more mid-sized clusters and significantly more \cdag{} edges (averaging $253$--$307$ versus \olaca{}'s $111$). Although fewer edges reduce the likelihood of cycles, this topology alone does not guarantee admissibility. Despite sharing \olaca{}'s exact dominant-cluster distribution, \emph{Rand} retains $278$ edges---exceeding most baselines. Thus, \olaca{} is admissible not because a single large cluster trivially absorbs random edges, but because GO successfully groups interacting co-participant genes into the same block.

\textbf{Missing annotations and unrecorded edges in GO account for all inadmissible instances.} 
Across all sizes and replicates, \olaca{} produced $52$ inadmissible instances encompassing $482$ edges within their SCCs (see Table~\ref{tab:all52} in Appendix~\ref{app:olaca_admiss} for full details). 
In every case, this was caused by missing data in GO, stemming from two distinct sources: missing gene annotations at the evaluated depth  (meaning GO could not identify the corresponding incoming or outgoing
edges), or regulatory edges were simply absent from GO knowledge base despite the presence of annotations. 
Of the $52$ instances, $9$ were strictly due to absent edge data between fully annotated nodes. Notably, all $9$ were 2-cluster loops involving the \emph{cytoplasmic translation} cluster. The remaining 43 instances exhibited a mix of annotated and unannotated genes.

\begin{table}[H]
\setlength{\abovecaptionskip}{0pt}
\setlength{\belowcaptionskip}{4pt}
\caption{ Percentage of inadmissible instances across different sizes $n$ for all methods, beside the number of \cdag{} edges that survive the coarsening and the \emph{shape} of the partitions being compared. G, L, R, W are \olaca{}, Lin, Resnik and Wang. \emph{largest} is the share of the $n$ genes in the biggest cluster and \emph{singletons} the share of the $k$ clusters holding one gene, both means over instances.}
\label{tab:premerge-admissibility}
\centering
{\scriptsize
\setlength{\tabcolsep}{3pt}
\begin{tabular}{r c ccccc ccccc cccc cccc}
\toprule
& & \multicolumn{5}{c}{inadmissible instances (\%)} & \multicolumn{5}{c}{C-DAG edges} & \multicolumn{4}{c}{largest (share of $n$)} & \multicolumn{4}{c}{singletons (share of $k$)} \\
\cmidrule(lr){3-7}\cmidrule(lr){8-12}\cmidrule(lr){13-16}\cmidrule(lr){17-20}
size & instance & G & L & R & W & Rand & G & L & R & W & Rand & G & L & R & W & G & L & R & W \\
\midrule
50 & 40 & \textbf{10.0} & 35.0 & 50.0 & 32.5 & 77.8 & \textbf{20} & 32 & 35 & 31 & 44 & \textbf{0.64} & 0.38 & 0.37 & 0.36 & \textbf{0.75} & 0.42 & 0.39 & 0.37 \\
100 & 38 & \textbf{13.2} & 44.7 & 52.6 & 52.6 & 80.8 & \textbf{40} & 90 & 94 & 91 & 108 & \textbf{0.65} & 0.31 & 0.30 & 0.30 & \textbf{0.79} & 0.44 & 0.42 & 0.41 \\
200 & 43 & \textbf{7.0} & 46.5 & 79.1 & 58.1 & 77.9 & \textbf{57} & 131 & 152 & 140 & 161 & \textbf{0.73} & 0.38 & 0.34 & 0.33 & \textbf{0.83} & 0.51 & 0.50 & 0.46 \\
400 & 43 & \textbf{7.0} & 81.4 & 95.3 & 88.4 & 80.2 & \textbf{110} & 261 & 311 & 261 & 297 & \textbf{0.73} & 0.28 & 0.27 & 0.28 & \textbf{0.87} & 0.41 & 0.38 & 0.49 \\
600 & 43 & \textbf{27.9} & 74.4 & 97.7 & 74.4 & 82.1 & \textbf{163} & 368 & 442 & 373 & 413 & \textbf{0.72} & 0.27 & 0.23 & 0.27 & \textbf{0.90} & 0.46 & 0.38 & 0.47 \\
800 & 40 & \textbf{30.0} & 87.5 & 100.0 & 77.5 & 82.5 & \textbf{161} & 396 & 487 & 380 & 411 & \textbf{0.76} & 0.25 & 0.22 & 0.26 & \textbf{0.91} & 0.38 & 0.33 & 0.39 \\
1000 & 41 & \textbf{31.7} & 85.4 & 97.6 & 92.7 & 84.9 & \textbf{222} & 499 & 609 & 479 & 496 & \textbf{0.75} & 0.24 & 0.21 & 0.23 & \textbf{0.92} & 0.40 & 0.35 & 0.39 \\
\midrule
All & 288 & \textbf{18.1} & 65.3 & 82.3 & 68.4 & 80.9 & \textbf{111} & 256 & 307 & 253 & 278 & \textbf{0.71} & 0.30 & 0.28 & 0.29 & \textbf{0.85} & 0.43 & 0.39 & 0.43 \\
\bottomrule
\end{tabular}}
\end{table}
\vspace{-2mm}

\subsection{Skeleton Recovery under Oracle Mode}
\textbf{Evaluation Metric.} We report the Structural Hamming Distance ($\mathrm{SHD}$), calculated over unordered cluster pairs rather than edges, normalized by $\binom{k}{2}$. We modify the standard SHD because \cloc{} cannot handle cycles, that is to say it cannot draw more than one edge between nodes. For a true cluster graph having a pair with edges into each other ($B \rightleftarrows A$), if \cloc{} assigns it an edge but leaves it unoriented it is not penalised. Whereas orientating it counted as an error since we consider committing to any direction in a cyclic pair wrong. Additionally, we count missed adjacencies (False Negatives), wrong adjacencies (False Positives) and misorientations with respect to the true cluster graph.

\textbf{\olaca{}'s partition gives the lowest SHD at almost every granularity.} Table~\ref{tab:pairwise-shd-oracle}
reports normalised $\mathrm{SHD}_{\text{pair}}$ for sizes 50 and 100, in $k$-bins
of width 10. \olaca{} is the lowest in 10 of the 11 bins, strictly so in 7 and
tied at the remaining three. \olaca{} is behind only at size 50 for $k = 41$--$50$. Since leaving an edge undirected costs nothing, what is left is
almost entirely invented adjacency: FP accounts for $92.0\%$ of \olaca{}'s
$\mathrm{SHD}$ and $87.4$--$88.6\%$ of the baseline's. (Full 
Table~\ref{tab:pshd-breakdown} in Appendix~\ref{app:oracle-results}).

\textbf{On \olaca{}'s partition \cloc{} almost never reverses an edge.} Among the directed edges of the true cluster graph that
\cloc{} orients, it points the wrong way on $0.2\%$ when it runs on \olaca{}'s
partition, against $10.0\%$, $11.3\%$ and $11.4\%$ on Lin's, Resnik's and
Wang's (Table~\ref{tab:pshd-edge-breakdown-all} in
Appendix~\ref{app:oracle-results}). In absolute terms that is $7$ reversed edges
out of the $1348$ \cloc{} directs on \olaca{}'s partition, against $479$, $646$
and $546$ on the baseline's. An orientation returned on \olaca{}'s partition is
therefore almost always correct. Restricted to the instances with zero False Positives, \cloc{} reverses $7$, $5$ and $8$ edges on \textbf{L}'s, \textbf{R}'s and \textbf{W}'s partitions and none on \olaca{}'s~(Table~\ref{tab:pshd-edge-breakdown-fpfree}).

\begin{table}[H]
\setlength{\abovecaptionskip}{0pt}
\setlength{\belowcaptionskip}{4pt}
\centering
\caption{SHD by $k$ bin, sizes 50 and 100. Mean$\pm$std over each bin. The remaining sizes see Table~\ref{tab:pairwise-shd-rest} in Appendix~\ref{app:oracle-results}.}
\label{tab:pairwise-shd-oracle}
{\scriptsize
\setlength{\tabcolsep}{3pt}
\begin{tabular}{r l r cccc @{\hspace{3pt}} r l r cccc}
\toprule
size & $k$ & inst. & G & L & R & W & size & $k$ & inst. & G & L & R & W \\
\midrule
50 & 1--10 & 18 & \textbf{.000{\tiny$\pm$.000}} & .002{\tiny$\pm$.010} & .001{\tiny$\pm$.005} & \textbf{.000{\tiny$\pm$.000}} & 100 & 1--10 & 14 & .000{\tiny$\pm$.000} & .000{\tiny$\pm$.000} & .000{\tiny$\pm$.000} & .000{\tiny$\pm$.000} \\
 & 11--20 & 9 & \textbf{.000{\tiny$\pm$.000}} & .023{\tiny$\pm$.047} & .038{\tiny$\pm$.092} & .019{\tiny$\pm$.039} &  & 11--20 & 4 & \textbf{.000{\tiny$\pm$.000}} & \textbf{.000{\tiny$\pm$.000}} & .004{\tiny$\pm$.007} & \textbf{.000{\tiny$\pm$.000}} \\
 & 21--30 & 6 & \textbf{.000{\tiny$\pm$.000}} & .154{\tiny$\pm$.129} & .217{\tiny$\pm$.097} & .147{\tiny$\pm$.147} &  & 21--30 & 3 & \textbf{.000{\tiny$\pm$.000}} & .077{\tiny$\pm$.104} & .178{\tiny$\pm$.176} & .056{\tiny$\pm$.045} \\
 & 31--40 & 6 & \textbf{.024{\tiny$\pm$.020}} & .187{\tiny$\pm$.093} & .211{\tiny$\pm$.080} & .220{\tiny$\pm$.108} &  & 31--40 & 2 & \textbf{.000{\tiny$\pm$.000}} & .084{\tiny$\pm$.077} & .166{\tiny$\pm$.086} & .111{\tiny$\pm$.101} \\
 & 41--50 & 1 & .103{\tiny$\pm$.000} & \textbf{.086{\tiny$\pm$.000}} & \textbf{.086{\tiny$\pm$.000}} & .128{\tiny$\pm$.000} &  & 41--50 & 4 & \textbf{.000{\tiny$\pm$.000}} & .294{\tiny$\pm$.085} & .333{\tiny$\pm$.089} & .292{\tiny$\pm$.072} \\
 &  &  &  &  &  &  &  & 51--60 & 5 & \textbf{.000{\tiny$\pm$.000}} & .315{\tiny$\pm$.075} & .343{\tiny$\pm$.081} & .331{\tiny$\pm$.075} \\
\bottomrule
\end{tabular}}
\end{table}

\vspace{-3mm}
\textbf{The MEC limit of each partition.} On an admissible partition with a perfect CI test \cloc{} is sound and
complete, so its output is the MEC of the true cluster graph (here the true cluster graph and \cdag{} are same). Therefore we filter instances that are \emph{admissible} and without any errors from imperfect CI test (no extra/missing adjacencies). \textbf{L} qualifies on
$95$ of the $176$ instances, \textbf{R} on $44$ and \textbf{W} on $84$, and \olaca{} qualifies on every one of these. Then we compare \olaca{} and each baseline pairwise (Table~\ref{tab:pshd-edge-breakdown-clean}, Appendix~\ref{app:oracle-results}). Against \textbf{L}, \olaca{} leaves $74.4\%$ of its edges undirected and \textbf{L}
$75.8\%$. Against \textbf{R} it leaves $93.4\%$ and \textbf{R} $100\%$, where no
edge of Resnik's cluster graph can be oriented at all. Against \textbf{W} it
leaves $76.6\%$ and \textbf{W} $72.8\%$, so there the tests reach further on
\textbf{W}'s partition. Each comparison runs on that baseline's own qualifying
instances.

\begin{figure}[H]
\centering
\includegraphics[width=\textwidth]{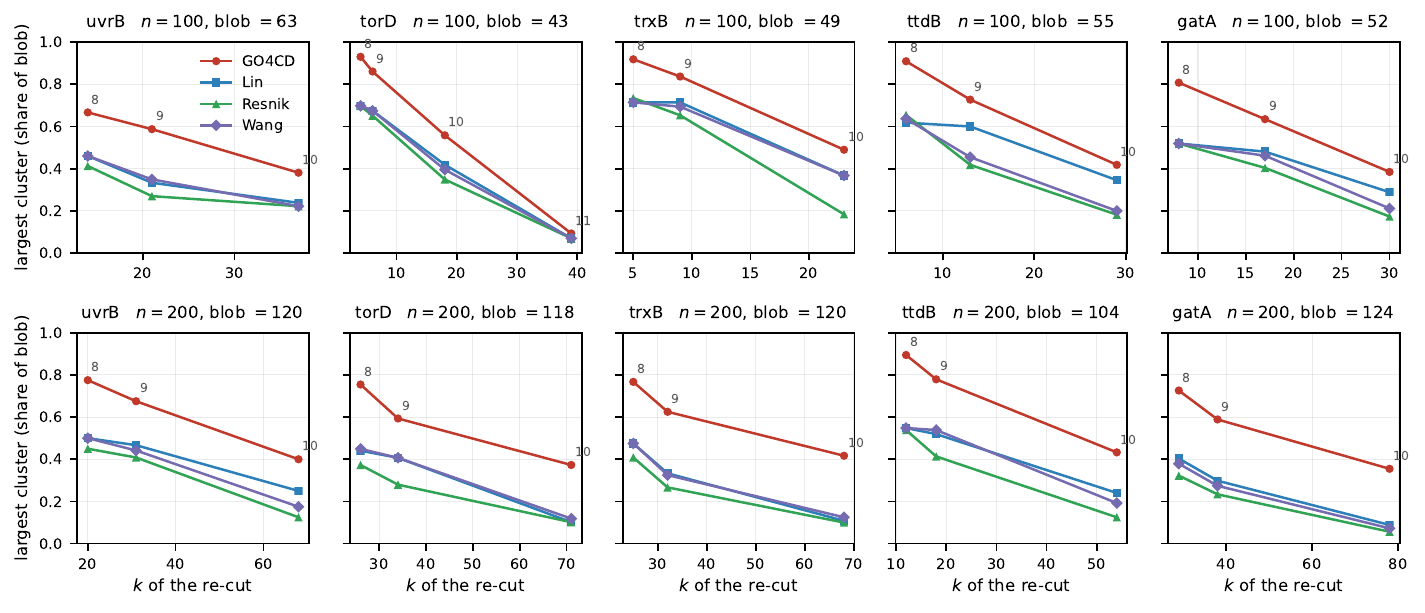}
\caption{Largest-cluster share against $k$ inside \olaca{}'s biggest cluster,
after that cluster is cut out as a network of its own and re-partitioned at
every deeper GO depth. Depths are marked inside the plot.}
\label{fig:blob-largest}
\end{figure}

\textbf{\olaca{} retains the dominant cluster.} \olaca{} typically yields one large cluster containing $71\%$ of the genes, alongside many singletons. While traversing deeper into the ontology fragments this large block, it also sacrifices admissibility. All \emph{inadmissible} instances occur at these greater depths (Tables~\ref{tab:kbin-shape}, \ref{tab:all52}, Appendix~\ref{app:admissibility}). When reclustering only the large block, \olaca{} is inadmissible on $32.3\%$ of instances, compared to $71.0\%$--$80.6\%$ for the baselines. Nevertheless, \olaca{} still retains the largest cluster, capturing $62\%$ of its genes against $35\%$ -- $41\%$ for baselines (Figure~\ref{fig:blob-largest}, Appendix~\ref{app:admissibility}).

\subsection{Causal Discovery on Gene Expression Data}
In order to check performance when \emph{admissibility} assumption is under strain on expression data, we filter for instances where at least one method's projected true cluster graph is cyclic which leaves us with $24$ instances. 
The filter is applied to all four methods together, with   \textbf{L}, \textbf{R}, and \textbf{W} being acyclic in $20$, $1$ and $19$ instances, respectively, while \olaca{} is acyclic in all cases. Secondly, we also compare the costs of independence tests for each baselines at the same cluster count. In Table~\ref{tab:data-macro-micro} we report macro ($\mathrm{P}_{mac}$) and micro precision ($\mathrm{P}_{mic}$), together with the separating phase (of \olaca{}) share of the worst-case test budget ($B$) each method spends. $B$ is same for all methods.

\textbf{\olaca{} recovers skeleton better.} In Table~\ref{tab:data-macro-micro} \olaca{} leads on both $\mathrm{P}_{mac}$ and $\mathrm{P}_{mic}$ with one exception being $\mathrm{P}_{mic}$ at size $50$ where $\textbf{R}$ is just $0.03$ points ahead. The gap is negligible at size $50$ and large at sizes $100$ and $200$. Recall is $1.00$ throughout, only \textbf{R} and \textbf{W} miss a single edge each, both at size $50$ (See Table~\ref{tab:data-all}, Appendix~\ref{app:data-results} in full breakdown).  
\vspace{-2mm}
\begin{table}[H]
\setlength{\abovecaptionskip}{-2pt}
\setlength{\belowcaptionskip}{4pt}
\centering
\caption{Precision over the learned cluster skeleton and the share of the worst-case CI budget $B$ spent, on the 24 instances. \emph{macro} averages the per-instance value, \emph{micro} is the ratio of sums.}
\label{tab:data-macro-micro}
{\scriptsize
\setlength{\tabcolsep}{3pt}
\begin{tabular}{l r l rrrr rrrr}
\toprule
& & & \multicolumn{4}{c}{precision $\uparrow$} & \multicolumn{4}{c}{sepset\,/\,$B$ $\downarrow$} \\
\cmidrule(lr){4-7}\cmidrule(lr){8-11}
size & inst & & G & L & R & W & G & L & R & W \\
\midrule
50 & 10 & \emph{macro} & \textbf{0.711} & 0.642 & 0.701 & 0.662 & \textbf{0.294} & 0.378 & 0.492 & 0.361 \\
 &  & \emph{micro} & 0.689 & 0.665 & \textbf{0.723} & 0.685 & \textbf{0.201} & 0.300 & 0.388 & 0.254 \\
\addlinespace[2pt]
100 & 3 & \emph{macro} & \textbf{0.796} & 0.446 & 0.420 & 0.293 & \textbf{0.360} & 0.670 & 0.757 & 0.810 \\
 &  & \emph{micro} & \textbf{0.750} & 0.444 & 0.392 & 0.289 & \textbf{0.311} & 0.531 & 0.661 & 0.676 \\
\addlinespace[2pt]
200 & 11 & \emph{macro} & \textbf{0.784} & 0.465 & 0.515 & 0.433 & \textbf{0.500} & 0.772 & 0.784 & 0.765 \\
 &  & \emph{micro} & \textbf{0.662} & 0.326 & 0.362 & 0.307 & \textbf{0.239} & 0.518 & 0.564 & 0.554 \\
\bottomrule
\end{tabular}}
\end{table}

\vspace{-3mm}

\textbf{\olaca{} costs fewer CI tests than the baselines.} Across all network sizes, \olaca{} consistently executes the fewest CI tests. Furthermore, its efficiency advantage scales with network size: compared to \olaca{}, the baselines issue $1.3$--$1.9\times$ more calls at 50 nodes, $1.7$--$2.2\times$ more at 100 nodes, and $2.2$--$2.4\times$ more at 200 nodes. On a per-instance basis, \olaca{} strictly requires the fewest tests on 20 out of 24 instances during the separation phase. While \olaca{}'s own testing cost naturally rises as networks grow (averaging shares of $0.29$, $0.36$, and $0.50$), it scales at a substantially slower rate than the baselines, causing the efficiency gap to widen continuously.

\section{Conclusion and Future Work}

Causal discovery over high-dimensional GRNs is
computationally intractable and difficult to interpret at the single-gene
resolution. Recent methods address this by abstracting or clustering genes, but operate under strict assumption of \emph{admissible}
partition provided to them. We bridge this gap with \olaca{}, an algorithm that uses knowledge from GO to abstract genes into biologically meaningful clusters along the ontology hierarchy. On
\emph{E. coli} subnetworks from RegulonDB, \olaca{} yielded admissible
partitions far more often than the baselines, and a size-preserving random
control tied this to the ontology's structure rather than the partition's
shape. We integrated \olaca{} with \cloc{} algorithm and reported its performance using oracle and multivariate CI test. For future work, we aim to use neuro-symbolic methods to learn partitions and then apply constraint based test for causal discovery on them.

\section*{AI Use Statement}

We used generative AI to assist with writing code, identifying relevant literature, and editing the manuscript for clarity. We manually reviewed and verified the AI-assisted content and take full responsibility for the final paper.
\section*{Ethics Statement}
We have read and adhered to the ICLR Code of Ethics. Our work is intended to support the study of relationships among gene modules. The inferred relationships are hypotheses that require further validation and should not be used on their own to guide clinical decisions. 

\section*{Reproducibility Statement}
The method section describes \olaca{}'s inputs and procedure. The experimental setup specifies the datasets, baselines, and evaluation metrics, while the appendix provides additional implementation details. Together, these descriptions support reproduction of the reported results.


\newpage
\appendix

\section{Baselines}\label{app:baseline}
\textbf{Term sets.} Each gene's term set is its participant annotations under the
same relation and evidence filters our own clustering applies, collected at any
depth rather than only at the depth being clustered.

\textbf{Information content.} $\mathrm{IC}(t) = -\log p(t)$, with $p(t)$ estimated
over every BP-annotated gene in the GAF rather than over the filtered set. The
common ancestor is the one of maximum IC (MICA), not the deepest, and the
ancestor closure runs over both \texttt{is\_a} and \texttt{part\_of}.

\textbf{Gene-pair score.} Best-match average, taken as the mean of the two
directional means,
\begin{equation*}
\mathrm{BMA}(g_1,g_2) \;=\;
\tfrac{1}{2}\left[\tfrac{1}{m}\sum_{a}\max_{b} s(a,b)
              + \tfrac{1}{n}\sum_{b}\max_{a} s(a,b)\right],
\end{equation*}
rather than the pooled
$\left[\sum_{a}\max_{b} s + \sum_{b}\max_{a} s\right]/(m+n)$.
The two differ whenever the genes carry different numbers of terms.

\textbf{Wang.} S-values propagate with weight $0.8$ across \texttt{is\_a} and
$0.6$ across \texttt{part\_of}, taking the maximum over multiple paths.

\textbf{Normalisation and clustering.} $S_{ij}/\sqrt{S_{ii}S_{jj}}$, distance
$D = 1 - S$, average linkage, cut at \olaca{}'s $k$ for that cell.

\textbf{Matched $k$.} \texttt{fcluster} with \texttt{criterion="maxclust"} returns
\emph{at most} $k$ clusters, and a baseline tree does not always resolve as many
clusters as \olaca{} does at that depth. Of the 396 (replicate, size, depth) cells
where all four methods are present, 88 have a $k$ mismatch; in every one of them
the baselines undershoot rather than overshoot, and all 88 sit at depth 10--13
against a median depth of 6 for the cells we keep. In 38 of the 88, all three
baselines are already at their own maximum $k$ for that subnetwork. We drop those
cells; dropping in addition the cells where $k = n$, the identity abstraction on
which all four methods agree trivially, leaves the 288 the tables report. Dropping them is the conservative
choice: at a smaller $k$ a baseline's C-DAG is coarser, and a coarser graph is
less likely to carry a cycle.

\section{Sampling subnetworks from RegulonDB}\label{app:samplesubnet}

\textbf{Source network and sampling.} We based our sampling on the \emph{E. coli} regulatory network from RegulonDB. After filtering for high-confidence edges and removing feedback loops, we were left with an acyclic network containing 1,781 genes and 3,416 edges. Because this network is disconnected, 1,699 of those genes form a single giant component, while the remaining 82 are scattered across 26 isolated fragments (none larger than 12 genes). If we started a sampling walk inside one of these small fragments, it would quickly hit a dead end, forcing the walk to artificially jump to unconnected genes. To prevent this, we seed every walk inside the giant component, guaranteeing that all extracted subnetworks remain fully connected.

\textbf{Subnetwork extraction.} To extract a subnetwork of size $n$, we start by
selecting a random seed gene from the giant component and grow the subnetwork
iteratively, adding one gene at a time. At each step, we identify all neighboring
genes---defined as any gene in the source network with a direct edge to the current
subnetwork. We count how many connections each neighbor shares with the subnetwork
and add the one with the highest count, breaking ties randomly. Edge directions are
ignored during this selection process because a gene is structurally attached
whether it regulates a gene inside the subnetwork or is regulated by one. However,
all original edge directions are preserved in the final extracted network.

\textbf{Comparison to previous methods.} Our overall procedure closely follows the
module extraction method of \citet{marbach2009generating}, with one key difference regarding
this selection criterion. Marbach et al.\ select the neighbor that maximizes the
subnetwork's modularity $Q$, defined as the actual number of internal edges minus
the expected number of edges in a randomized network with an identical degree
sequence:
\begin{equation}
    Q = \frac{1}{4m} s^\top B s, \quad B_{ij} = A_{ij} - \frac{k_i k_j}{2m}
\end{equation}
where $A_{ij}$ is the edge count between genes $i$ and $j$, $k_i$ is the degree of
gene $i$, $m$ is the total number of edges, and $s$ is the subnetwork indicator
vector ($s_i = 1$ if gene $i$ is included, $-1$ otherwise).

Instead of using modularity, our rule relies solely on the raw number of edges
connecting the neighbor to the subnetwork. The expected edge penalty
($\frac{k_i k_j}{2m}$) in $Q$ disproportionately penalizes high-degree genes; if a
gene regulates a large portion of the network, it is statistically expected to have
edges pointing into any subnetwork, meaning those edges count for very little under
the modularity metric. By dropping the penalty and counting all edges fully, our
method incorporates these crucial global regulators into the subnetwork much
earlier.

\textbf{Growth runs and replicates.} For each replicate, we perform a single growth
run up to 1,000 genes. Smaller subnetwork sizes are taken as prefixes of this run,
meaning the 50-gene network is nested within the 100-gene network, which is nested
within the 200-gene network, and so forth. For each target size, we extract the
induced subgraph from the source network, retaining only the existing edges between
the selected genes. Because an induced subgraph of an acyclic network inherently
remains acyclic, we do not need to perform any additional cycle-breaking.

Ultimately, we generated 35 subnetworks by extracting seven different sizes
(ranging from 50 to 1,000 genes) across five replicates seeded with the genes
\emph{uvrB}, \emph{torD}, \emph{trxB}, \emph{ttdB}, and \emph{gatA}. The resulting
edge counts range from 128--212 for the 50-gene networks to 2,612--2,623 for the
1,000-gene networks. Finally, because we simulate expression separately, we use
this entire procedure strictly to establish the network topology.

\section{Data Generation}\label{app:datagen}

\textbf{Why we chose CausalRegNet.} We generated all expression data with
CausalRegNet \citep{kovavcevic2024simulation}, an explicit structural causal model for
single-cell counts. CausalRegNet draws each gene once from its parents and its
own independent noise, and it never integrates a trajectory over time, so its
sampling distribution is Markov to the input DAG by construction. Simulators
like SERGIO~\citep{sergio} work the other way round, since they evolve a stochastic
differential equation to a pseudo steady state and then sample from it, and the
shared dynamics introduce dependencies the DAG never licensed. We confirmed this
difference on the 21-gene \emph{E.\ coli} SOS network, where SERGIO's data was
not Markov to its own GRN. On the same network all six conditional independence
tests agreed with the graph once we drew at least 20,000 cells from
CausalRegNet, which is why we generate well above that count.

\textbf{The generative model.} CausalRegNet visits the genes in topological
order and draws each one from a Negative Binomial. For gene $j$ with parents
$\Pa(j)$, the regulators are first pooled into a single score,
\begin{equation}
r_j = \sum_{i \in \Pa(j)} W_{ij}\,\frac{X_i}{\mu_i} + a_j,
\end{equation}
where $W_{ij}$ is the weight of the edge $i \to j$, $\mu_i$ is gene $i$'s
observational mean and $a_j$ is a per-gene regulatory constant. This score then
passes through a sigmoid, which makes a gene's dependence on its parents
nonlinear and bounds how far the regulators can move the target,
\begin{equation}
f_j(r_j) = \frac{\alpha_j}{1 + \exp\!\left(-\gamma_j (r_j + b_j)\right)},
\end{equation}
and gene $j$ is drawn with mean $m_j = \mu_j f_j(r_j)$ and variance
$m_j(1 + m_j/\theta_j)$. A gene with no parents is drawn with mean $\mu_j$
instead. The ceiling $\alpha_j$ and the floor $\beta_j$ bound the regulatory
effect, and CausalRegNet calibrates the steepness $\gamma_j$ and the offset
$b_j$ per gene from those two bounds, the incoming weights and $a_j$.

\textbf{Per-gene parameters.} We kept the defaults from the CausalRegNet paper's
own synthetic benchmark, so we did not tune the simulator to our networks. We
drew $\mu_j \sim U(1,5)$ and $\theta_j \sim U(2,15)$ independently for each
gene, and set $\alpha_j = 2$, $\beta_j = 0.1$ and $a_j = 10$ for every gene.

\textbf{Edge signs and weights.} Every edge needs a signed weight, and we took
the sign from RegulonDB's~\citep{salgado2023regulondb} own function annotation rather than assigning it
ourselves. An edge marked as an activator gets a positive sign and one marked as
a repressor gets a negative sign. We then drew the magnitude independently per
edge from $U(1,3)$.

We drew the magnitude randomly to avoid an artifact we had already run into. If
every activator carries one fixed magnitude and every repressor carries its
negative, then a gene with an equal number of activators and repressors receives
exactly zero net input, so its regulators leave no trace in the data. This case
is common in RegulonDB rather than rare. Across our 35 subnetworks, 2,087 of the
14,732 regulated genes (14.2\%) have exactly as many activators as repressors,
counting a gene once for each subnetwork it appears in. On the 26-gene
SOS-extended network this cancellation was the confirmed cause of a cluster edge
that we could not recover. Drawing continuous magnitudes makes an exact
cancellation a probability-zero event, and it does so without us tuning any
individual gene.

\textbf{Edges with no usable sign.} RegulonDB records some edges with
conflicting activation and repression evidence, and a few with no sign character
at all. We assign each of these a random sign from the same random stream.
Across the 35 subnetworks this affects 4,018 of 49,154 edges (8.2\%), of which
16 are unsigned and the rest carry dual evidence. These edges keep their
position in the graph but not their recorded direction of effect, so our
simulated data follows the graph we score against while departing from
RegulonDB's regulation semantics on this small fraction of edges.

\textbf{What we generated.} We simulated 100,000 cells for each of the 35 subnetworks. Every run starts from seed 2026, so a run reproduces exactly, and the datasets differ because the networks do, not because the seeds do. Each dataset is generated from the same acyclic gene-level DAG that
we coarsen to obtain the ground-truth \cdag{}, so the data and the ground truth
come from one structure rather than two. We check acyclicity before simulating
and stop if the check fails. All randomness comes from a single seeded stream
consumed in a fixed order, first the signs of the ambiguous edges, then the
weight magnitudes, then $\mu$, then $\theta$, and finally the simulation draws,
so a given seed reproduces a dataset end to end.

\section{Additional Results}

\subsection{Admissibility Results \olaca{} vs. Baselines}
\label{app:admissibility}

\subsubsection{Admissibility in \olaca{}}\label{app:olaca_admiss}

\begin{figure}[H]
\setlength{\abovecaptionskip}{-15pt}
\setlength{\belowcaptionskip}{-15pt}
\centering
\includegraphics[width=\textwidth]{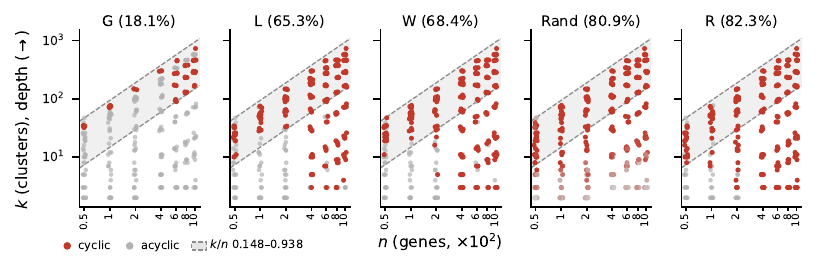}
\caption{Where the inadmissible cells sit in (n, k). Inadmissible cells for different sizes for all the methods. Each of the 308 matched-$k$ cells is a point. Shading is $0.148 \leq k/n \leq 1$, below 0.148 (lower bound) \olaca{} is never cyclic.}
\label{fig:inadmissible}
\end{figure}

\textbf{Which blocks we re-cut.} The $288$ instances sit inside $35$ networks,
and within a network the blocks at successive depths are nested, so we take one
block per network rather than one per instance. We take it at the shallowest
depth whose largest cluster holds at most $70\%$ of the $n$ genes: above that
the block is close to the whole network and re-clustering it asks nothing. That
rule lands on depth $7$ in all ten networks of sizes $100$ and $200$, giving
blocks of $43$--$124$ genes and $119$--$452$ induced edges. We exclude size $50$,
where the rule picks a different depth in every replicate and the blocks split
into two or three clusters, which leaves nothing to match on. Each block is then
cut out as a network of its own and re-partitioned at every deeper depth, with
the baselines re-cut from a fresh average-linkage tree over the block's genes at
\olaca{}'s matched $k$. This gives $50$ block-depth pairs, of which $31$ have all
four methods at the same $k$; the remaining $19$ fall to the same constraint as
the main ladder, the baselines landing below \olaca{}'s $k$ and never above it.
We never condition on \olaca{} being admissible, since the block is there in
every instance regardless.

\begin{table}[H]
\setlength{\abovecaptionskip}{5pt}
\setlength{\belowcaptionskip}{8pt}
\centering
\caption{\small How the four methods cut the same genes, within each network size and $k$ bin. \emph{largest} is the share of the $n$ genes sitting in the biggest cluster, \emph{singletons} the share of the $k$ clusters holding one gene; both are means over the instances in the bin. $k$ is identical across the four methods within a instance, so a row holds granularity fixed. Bins with no instances are omitted, as are instances with $k = n$. G, L, R, W are \olaca{}, Lin, Resnik and Wang.}
\label{tab:kbin-shape}
{\scriptsize
\setlength{\tabcolsep}{1.2pt}
\begin{tabular}{l r cccc cccc @{\hskip 3pt} l r cccc cccc}
\toprule
\multicolumn{10}{c}{size 50} & \multicolumn{10}{c}{size 100} \\
& & \multicolumn{4}{c}{largest (share of $n$)} & \multicolumn{4}{c}{singletons (share of $k$)} & & & \multicolumn{4}{c}{largest (share of $n$)} & \multicolumn{4}{c}{singletons (share of $k$)} \\
\multicolumn{1}{c}{$k$} & instances &G& L & R & W &G& L & R & W & \multicolumn{1}{c}{$k$} & instances &G& L & R & W &G& L & R & W \\
\midrule
1--10 & 18 & \textbf{0.88} & 0.58 & 0.56 & 0.51 & \textbf{0.62} & 0.21 & 0.17 & 0.11 & 1--20 & 18 & \textbf{0.94} & 0.49 & 0.49 & 0.49 & \textbf{0.67} & 0.16 & 0.14 & 0.11 \\
11--20 & 9 & \textbf{0.59} & 0.32 & 0.28 & 0.32 & \textbf{0.77} & 0.45 & 0.40 & 0.45 & 21--40 & 5 & \textbf{0.52} & 0.26 & 0.23 & 0.24 & \textbf{0.81} & 0.56 & 0.49 & 0.56 \\
21--30 & 6 & \textbf{0.44} & 0.21 & 0.23 & 0.23 & \textbf{0.89} & 0.62 & 0.59 & 0.61 & 41--60 & 9 & \textbf{0.42} & 0.14 & 0.13 & 0.13 & \textbf{0.91} & 0.69 & 0.70 & 0.68 \\
31--40 & 6 & \textbf{0.28} & 0.12 & 0.12 & 0.12 & \textbf{0.92} & 0.74 & 0.76 & 0.73 & 61--80 & 6 & \textbf{0.25} & 0.07 & 0.07 & 0.06 & \textbf{0.94} & 0.79 & 0.79 & 0.78 \\
41--50 & 1 & 0.06 & 0.06 & 0.06 & 0.06 & 0.96 & 0.96 & 0.96 & 0.96 & & & & & & & & & &  \\
\midrule
\multicolumn{10}{c}{size 200} & \multicolumn{10}{c}{size 400} \\
& & \multicolumn{4}{c}{largest (share of $n$)} & \multicolumn{4}{c}{singletons (share of $k$)} & & & \multicolumn{4}{c}{largest (share of $n$)} & \multicolumn{4}{c}{singletons (share of $k$)} \\
\multicolumn{1}{c}{$k$} & instances &G& L & R & W & G & L & R & W & \multicolumn{1}{c}{$k$} & instances & G & L & R & W & G & L & R & W \\
\midrule
1--40 & 25 & \textbf{0.94} & 0.55 & 0.52 & 0.49 & \textbf{0.76} & 0.36 & 0.35 & 0.31 & 1--80 & 25 & \textbf{0.94} & 0.41 & 0.41 & 0.42 & \textbf{0.83} & 0.24 & 0.21 & 0.36 \\
41--80 & 5 & \textbf{0.59} & 0.22 & 0.19 & 0.21 & \textbf{0.92} & 0.59 & 0.55 & 0.57 & 81--160 & 5 & \textbf{0.65} & 0.18 & 0.11 & 0.17 & \textbf{0.93} & 0.50 & 0.40 & 0.51 \\
81--120 & 10 & \textbf{0.42} & 0.14 & 0.08 & 0.10 & \textbf{0.94} & 0.74 & 0.73 & 0.70 & 161--240 & 10 & \textbf{0.39} & 0.09 & 0.05 & 0.08 & \textbf{0.93} & 0.68 & 0.67 & 0.68 \\
121--160 & 3 & \textbf{0.23} & 0.03 & 0.03 & 0.04 & \textbf{0.96} & 0.80 & 0.80 & 0.80 & 241--320 & 3 & \textbf{0.20} & 0.02 & 0.02 & 0.02 & \textbf{0.97} & 0.83 & 0.83 & 0.83 \\
\midrule
\multicolumn{10}{c}{size 600} & \multicolumn{10}{c}{size 800} \\
& & \multicolumn{4}{c}{largest (share of $n$)} & \multicolumn{4}{c}{singletons (share of $k$)} & & & \multicolumn{4}{c}{largest (share of $n$)} & \multicolumn{4}{c}{singletons (share of $k$)} \\
\multicolumn{1}{c}{$k$} & instances & G & L & R & W & G & L & R & W & \multicolumn{1}{c}{$k$} & instances & G & L & R & W & G & L & R & W \\
\midrule
1--120 & 25 & \textbf{0.94} & 0.41 & 0.35 & 0.40 & \textbf{0.87} & 0.31 & 0.21 & 0.33 & 1--160 & 25 & \textbf{0.94} & 0.35 & 0.32 & 0.36 & \textbf{0.88} & 0.24 & 0.18 & 0.25 \\
121--240 & 5 & \textbf{0.65} & 0.12 & 0.10 & 0.12 & \textbf{0.94} & 0.49 & 0.39 & 0.48 & 161--320 & 5 & \textbf{0.66} & 0.10 & 0.07 & 0.11 & \textbf{0.95} & 0.47 & 0.44 & 0.50 \\
241--360 & 10 & \textbf{0.38} & 0.08 & 0.05 & 0.07 & \textbf{0.93} & 0.68 & 0.66 & 0.68 & 321--480 & 10 & \textbf{0.37} & 0.07 & 0.04 & 0.06 & \textbf{0.94} & 0.69 & 0.67 & 0.68 \\
361--480 & 3 & \textbf{0.17} & 0.02 & 0.02 & 0.02 & \textbf{0.97} & 0.84 & 0.83 & 0.83 & & & & & & & & & &  \\
\midrule
\multicolumn{10}{c}{size 1000} & \multicolumn{10}{c}{averaged over all instances for all sizes} \\
& & \multicolumn{4}{c}{largest (share of $n$)} & \multicolumn{4}{c}{singletons (share of $k$)} & & & \multicolumn{4}{c}{largest (share of $n$)} & \multicolumn{4}{c}{singletons (share of $k$)} \\
\multicolumn{1}{c}{$k$} & instances & G & L & R & W & G & L & R & W & \multicolumn{1}{c}{$k$} & instances & G & L & R & W & G & L & R & W \\
\midrule
1--200 & 25 & \textbf{0.94} & 0.35 & 0.31 & 0.34 & \textbf{0.89} & 0.26 & 0.19 & 0.23 & all & 288 & \textbf{0.71} & 0.30 & 0.28 & 0.29 & \textbf{0.85} & 0.43 & 0.39 & 0.43 \\
201--400 & 5 & \textbf{0.67} & 0.09 & 0.07 & 0.10 & \textbf{0.96} & 0.47 & 0.44 & 0.49 & & & & & & & & & &  \\
401--600 & 10 & \textbf{0.38} & 0.06 & 0.03 & 0.06 & \textbf{0.95} & 0.68 & 0.67 & 0.68 & & & & & & & & & &  \\
601--800 & 1 & \textbf{0.18} & 0.02 & 0.02 & 0.02 & \textbf{0.97} & 0.83 & 0.83 & 0.82 & & & & & & & & & &  \\
\bottomrule
\end{tabular}}
\end{table}

\begin{table}[H]
\setlength{\abovecaptionskip}{5pt}
\setlength{\belowcaptionskip}{8pt}
\caption{All 52 instances where \olaca{}'s C-DAG is cyclic, in four column blocks.
$d$ and $k$ are the cyclic depth and its cluster count; $d'$ and $k'$ are the
deepest shallower depth whose C-DAG is acyclic and the cluster count there.}
\label{tab:all52}
\centering
{\scriptsize
\setlength{\tabcolsep}{2pt}
\begin{tabular}{rlrrrr @{\hspace{0.5em}} rlrrrr @{\hspace{0.5em}} rlrrrr @{\hspace{0.5em}} rlrrrr}
\toprule
size & rep & $d$ & $k$ & $d'$ & $k'$ & size & rep & $d$ & $k$ & $d'$ & $k'$ & size & rep & $d$ & $k$ & $d'$ & $k'$ & size & rep & $d$ & $k$ & $d'$ & $k'$ \\
\midrule
50 & torD & 10 & 35 & 9 & 26 & 400 & trxB & 10 & 301 & 9 & 219 & 600 & ttdB & 10 & 452 & 9 & 338 & 1000 & gatA & 7 & 292 & 6 & 166 \\
50 & trxB & 10 & 34 & 9 & 24 & 400 & uvrB & 10 & 296 & 9 & 212 & 800 & torD & 6 & 127 & 5 & 54 & 1000 & gatA & 8 & 460 & 6 & 166 \\
50 & ttdB & 10 & 32 & 9 & 19 & 600 & gatA & 10 & 446 & 9 & 334 & 800 & torD & 7 & 232 & 5 & 54 & 1000 & torD & 6 & 164 & 5 & 72 \\
50 & uvrB & 10 & 34 & 9 & 22 & 600 & torD & 6 & 89 & 5 & 40 & 800 & torD & 8 & 371 & 5 & 54 & 1000 & torD & 7 & 283 & 5 & 72 \\
100 & gatA & 10 & 77 & 9 & 61 & 600 & torD & 7 & 168 & 5 & 40 & 800 & trxB & 6 & 131 & 5 & 56 & 1000 & torD & 8 & 451 & 5 & 72 \\
100 & torD & 10 & 72 & 9 & 57 & 600 & torD & 8 & 270 & 5 & 40 & 800 & trxB & 7 & 236 & 5 & 56 & 1000 & trxB & 7 & 290 & 6 & 165 \\
100 & trxB & 10 & 70 & 9 & 53 & 600 & trxB & 6 & 93 & 5 & 40 & 800 & trxB & 8 & 378 & 5 & 56 & 1000 & trxB & 8 & 459 & 6 & 165 \\
100 & ttdB & 10 & 74 & 9 & 55 & 600 & trxB & 7 & 172 & 5 & 40 & 800 & ttdB & 6 & 130 & 5 & 57 & 1000 & trxB & 10 & 729 & 9 & 568 \\
100 & uvrB & 10 & 74 & 9 & 54 & 600 & trxB & 8 & 278 & 5 & 40 & 800 & ttdB & 7 & 230 & 5 & 57 & 1000 & ttdB & 6 & 170 & 5 & 82 \\
200 & gatA & 10 & 148 & 9 & 106 & 600 & trxB & 10 & 452 & 9 & 342 & 800 & ttdB & 8 & 370 & 5 & 57 & 1000 & ttdB & 7 & 295 & 5 & 82 \\
200 & torD & 10 & 147 & 9 & 109 & 600 & ttdB & 6 & 97 & 5 & 49 & 800 & uvrB & 6 & 126 & 5 & 58 & 1000 & ttdB & 8 & 466 & 5 & 82 \\
200 & uvrB & 10 & 143 & 9 & 102 & 600 & ttdB & 7 & 171 & 5 & 49 & 800 & uvrB & 7 & 232 & 5 & 58 & 1000 & uvrB & 7 & 280 & 6 & 158 \\
400 & torD & 10 & 302 & 9 & 219 & 600 & ttdB & 8 & 273 & 5 & 49 & 800 & uvrB & 8 & 375 & 5 & 58 & 1000 & uvrB & 8 & 451 & 6 & 158 \\
\bottomrule
\end{tabular}}
\end{table}

\subsection{Oracle Results}\label{app:oracle-results}

\begin{table}[H]
\setlength{\abovecaptionskip}{5pt}
\setlength{\belowcaptionskip}{8pt}
\centering
\caption{SHD by $k$ bin, sizes 200, 400 and 600, 800, 1000. Mean$\pm$std over each bin.}
\label{tab:pairwise-shd-rest}
{\scriptsize
\setlength{\tabcolsep}{1pt}
\begin{tabular}{r l cccc @{\hspace{3pt}} r l cccc}
\toprule
size & $k$ & G & L & R & W & size & $k$ & G & L & R & W \\
\midrule
200 & 1--10 & .000{\tiny$\pm$.000} & .000{\tiny$\pm$.000} & .000{\tiny$\pm$.000} & .000{\tiny$\pm$.000} & 600 & 1--10 & .000{\tiny$\pm$.000} & .000{\tiny$\pm$.000} & .000{\tiny$\pm$.000} & .000{\tiny$\pm$.000} \\
 & 11--20 & \textbf{.000{\tiny$\pm$.000}} & \textbf{.000{\tiny$\pm$.000}} & .005{\tiny$\pm$.010} & .002{\tiny$\pm$.003} &  & 11--20 & \textbf{.000{\tiny$\pm$.000}} & \textbf{.000{\tiny$\pm$.000}} & .011{\tiny$\pm$.012} & \textbf{.000{\tiny$\pm$.000}} \\
 & 21--30 & \textbf{.000{\tiny$\pm$.000}} & \textbf{.000{\tiny$\pm$.000}} & .006{\tiny$\pm$.006} & .001{\tiny$\pm$.001} &  & 31--40 & \textbf{.000{\tiny$\pm$.000}} & .001{\tiny$\pm$.001} & .011{\tiny$\pm$.008} & .000{\tiny$\pm$.001} \\
 & 31--40 & \textbf{.000{\tiny$\pm$.000}} & \textbf{.000{\tiny$\pm$.000}} & .001{\tiny$\pm$.001} & .003{\tiny$\pm$.001} &  & 41--50 & \textbf{.000{\tiny$\pm$.000}} & \textbf{.000{\tiny$\pm$.000}} & .013{\tiny$\pm$.005} & \textbf{.000{\tiny$\pm$.000}} \\
\cmidrule(lr){7-12}
 & 51--60 & \textbf{.000{\tiny$\pm$.000}} & .004{\tiny$\pm$.002} & .066{\tiny$\pm$.012} & .004{\tiny$\pm$.003} & 800 & 1--10 & \textbf{.000{\tiny$\pm$.000}} & \textbf{.000{\tiny$\pm$.000}} & .006{\tiny$\pm$.013} & \textbf{.000{\tiny$\pm$.000}} \\
\cmidrule(lr){1-6}
400 & 1--10 & .000{\tiny$\pm$.000} & .000{\tiny$\pm$.000} & .000{\tiny$\pm$.000} & .000{\tiny$\pm$.000} &  & 11--20 & \textbf{.000{\tiny$\pm$.000}} & \textbf{.000{\tiny$\pm$.000}} & .011{\tiny$\pm$.004} & \textbf{.000{\tiny$\pm$.000}} \\
 & 11--20 & .000{\tiny$\pm$.000} & .000{\tiny$\pm$.000} & .000{\tiny$\pm$.000} & .000{\tiny$\pm$.000} &  & 51--60 & \textbf{.000{\tiny$\pm$.000}} & .001{\tiny$\pm$.001} & .010{\tiny$\pm$.006} & .000{\tiny$\pm$.000} \\
\cmidrule(lr){7-12}
 & 21--30 & \textbf{.000{\tiny$\pm$.000}} & \textbf{.000{\tiny$\pm$.000}} & .014{\tiny$\pm$.023} & \textbf{.000{\tiny$\pm$.000}} & 1000 & 1--10 & .000{\tiny$\pm$.000} & .000{\tiny$\pm$.000} & .000{\tiny$\pm$.000} & .000{\tiny$\pm$.000} \\
 & 31--40 & .000{\tiny$\pm$.000} & .000{\tiny$\pm$.000} & .000{\tiny$\pm$.000} & .000{\tiny$\pm$.000} &  & 11--20 & \textbf{.000{\tiny$\pm$.000}} & \textbf{.000{\tiny$\pm$.000}} & .023{\tiny$\pm$.009} & \textbf{.000{\tiny$\pm$.000}} \\
 & 51--60 & \textbf{.000{\tiny$\pm$.000}} & \textbf{.000{\tiny$\pm$.000}} & .033{\tiny$\pm$.011} & .001{\tiny$\pm$.001} &  & 21--30 & \textbf{.000{\tiny$\pm$.000}} & .001{\tiny$\pm$.001} & .035{\tiny$\pm$.012} & .001{\tiny$\pm$.002} \\
\bottomrule
\end{tabular}}
\end{table}

\begin{table}[H]
\setlength{\abovecaptionskip}{5pt}
\setlength{\belowcaptionskip}{8pt}
\centering
\caption{Share of each method's SHD coming from false positives, reversed directions and 2-cycles, summed over the 176 instances. Edge is never missed so FN is always zero.}
\label{tab:pshd-breakdown}
{\scriptsize
\setlength{\tabcolsep}{6pt}
\begin{tabular}{l r @{\hspace{10pt}} r r r}
\toprule
& & \multicolumn{3}{c}{share of charge (\%)} \\
\cmidrule(lr){3-5}
method & SHD/instance & FP & reversed & 2-cycle \\
\midrule
\olaca{} & 1.13 & 92.0 & 3.5 & 4.5 \\
Lin & 27.31 & 88.6 & 10.0 & 1.4 \\
Resnik & 35.05 & 87.4 & 10.5 & 2.1 \\
Wang & 28.91 & 87.8 & 10.7 & 1.5 \\
\bottomrule
\end{tabular}}
\end{table}

\begin{table}[H]
\setlength{\abovecaptionskip}{5pt}
\setlength{\belowcaptionskip}{8pt}
\centering
\caption{Breakdown of the directed edges in the true cluster graph, for all instances. Correct, undirected and reversed refer to \cloc{}'s orientation of each edge.}
\label{tab:pshd-edge-breakdown-all}
{\scriptsize
\setlength{\tabcolsep}{8pt}
\begin{tabular}{l r r r r}
\toprule
method & edges & correct & undirected & reversed \\
\midrule
\olaca{} & 3058 & 1341 (43.9\%) & 1710 (55.9\%) & 7 (0.2\%) \\
Lin & 4780 & 2687 (56.2\%) & 1614 (33.8\%) & 479 (10.0\%) \\
Resnik & 5726 & 3071 (53.6\%) & 2009 (35.1\%) & 646 (11.3\%) \\
Wang & 4791 & 2734 (57.1\%) & 1511 (31.5\%) & 546 (11.4\%) \\
\bottomrule
\end{tabular}}
\end{table}

\textbf{Misorientations stem from false positives.} Filtering for instances with
zero False Positives (FPs) leaves $147$, $117$, and $147$ instances for pairwise
comparison between \olaca{} and \textbf{L}, \textbf{R}, and \textbf{W},
respectively. In these cases, \cloc{} reverses no edge at all on \olaca{}'s
partition, against $7$, $5$ and $8$ edges on \textbf{L}'s, \textbf{R}'s and
\textbf{W}'s (see Table~\ref{tab:pshd-edge-breakdown-fpfree} in
Appendix~\ref{app:oracle-results}). Over all $176$ instances the same three
partitions carry $479$, $646$ and $546$ reversals, so removing the invented
adjacencies removes almost every misorientation with them. Mechanistically, a
false edge shields a triple that was previously open, preventing the \cloc{}
orientation rules that fix edge directions from firing.

\begin{table}[H]
\setlength{\abovecaptionskip}{5pt}
\setlength{\belowcaptionskip}{8pt}
\centering
\caption{Breakdown of \cloc{}'s output on instances with no FP.}
\label{tab:pshd-edge-breakdown-fpfree}
{\scriptsize
\setlength{\tabcolsep}{8pt}
\begin{tabular}{l r r r r}
\toprule
method & edges & correct & undirected & reversed \\
\midrule
\multicolumn{5}{l}{\emph{\olaca{} vs Lin --- 147 instances}} \\
\olaca{} & 1808 & 453 (25.1\%) & 1355 (74.9\%) & 0 (0.0\%) \\
Lin & 1965 & 528 (26.9\%) & 1430 (72.8\%) & 7 (0.4\%) \\
\midrule
\multicolumn{5}{l}{\emph{\olaca{} vs Resnik --- 117 instances}} \\
\olaca{} & 832 & 169 (20.3\%) & 663 (79.7\%) & 0 (0.0\%) \\
Resnik & 998 & 167 (16.7\%) & 826 (82.8\%) & 5 (0.5\%) \\
\midrule
\multicolumn{5}{l}{\emph{\olaca{} vs Wang --- 147 instances}} \\
\olaca{} & 1784 & 477 (26.7\%) & 1307 (73.3\%) & 0 (0.0\%) \\
Wang & 1945 & 656 (33.7\%) & 1281 (65.9\%) & 8 (0.4\%) \\
\bottomrule
\end{tabular}}
\end{table}

\begin{table}[H]
\setlength{\abovecaptionskip}{5pt}
\setlength{\belowcaptionskip}{8pt}
\centering
\caption{Breakdown of the true cluster graph's directed edges on the admissible, FP $= 0$ instances. Every oriented edge here is correct, since \cloc{} is sound and complete in this regime.}
\label{tab:pshd-edge-breakdown-clean}
{\scriptsize
\setlength{\tabcolsep}{8pt}
\begin{tabular}{l r r r}
\toprule
method & edges & oriented & undirected \\
\midrule
\multicolumn{4}{l}{\emph{\olaca{} vs Lin --- 95 instances}} \\
\olaca{} & 808 & 207 (25.6\%) & 601 (74.4\%) \\
Lin & 887 & 215 (24.2\%) & 672 (75.8\%) \\
\midrule
\multicolumn{4}{l}{\emph{\olaca{} vs Resnik --- 44 instances}} \\
\olaca{} & 151 & 10 (6.6\%) & 141 (93.4\%) \\
Resnik & 159 & 0 (0.0\%) & 159 (100.0\%) \\
\midrule
\multicolumn{4}{l}{\emph{\olaca{} vs Wang --- 84 instances}} \\
\olaca{} & 624 & 146 (23.4\%) & 478 (76.6\%) \\
Wang & 679 & 185 (27.2\%) & 494 (72.8\%) \\
\bottomrule
\end{tabular}}
\end{table}

\subsection{Gene Expression Results}\label{app:data-results}

\begin{table}[H]
\setlength{\abovecaptionskip}{5pt}
\setlength{\belowcaptionskip}{8pt}
\centering
\caption{Per-instance results on the 24 instances at $k \le 20$ where at least one method's projected true C-DAG carries a cycle, named in the \emph{cyc} column. $|E|$ is that graph's edge count, $g_*$ the largest cluster's gene count, and precision and recall are over the learned skeleton. G\,/\,L\,/\,R\,/\,W = \olaca{}\,/\,Lin\,/\,Resnik\,/\,Wang. Each size block ends with \emph{macro} and \emph{micro} rows: the mean of the per-instance values and the ratio of sums, the latter undefined for $g_*$.}
\label{tab:data-all}
{\scriptsize
\setlength{\tabcolsep}{2.5pt}
\begin{tabular}{ccc r l rrrr rrrr rrrr rrrr }
\toprule
& & & & & \multicolumn{4}{c}{$|E|$} & \multicolumn{4}{c}{$g_*$} & \multicolumn{4}{c}{precision} & \multicolumn{4}{c}{recall} \\
\cmidrule(lr){6-9}\cmidrule(lr){10-13}\cmidrule(lr){14-17}\cmidrule(lr){18-21}
size & rep & depth & $k$ & cyc & G & L & R & W & G & L & R & W & G & L & R & W & G & L & R & W \\
\midrule
50 & uvrB & 8 & 16 & R & 15 & 22 & 25 & 22 & 29 & 16 & 14 & 16 & 0.88 & 0.85 & 0.89 & 0.88 & 1.00 & 1.00 & 1.00 & 1.00 \\
50 & torD & 6 & 14 & R & 13 & 13 & 24 & 13 & 25 & 16 & 16 & 16 & 0.72 & 0.62 & 0.75 & 0.62 & 1.00 & 1.00 & 1.00 & 1.00 \\
50 & torD & 7 & 19 & L R W & 18 & 19 & 33 & 19 & 23 & 15 & 15 & 15 & 0.67 & 0.66 & 0.89 & 0.68 & 1.00 & 1.00 & 1.00 & 1.00 \\
50 & trxB & 5 & 8 & R & 7 & 7 & 9 & 7 & 35 & 22 & 23 & 21 & 0.54 & 0.47 & 0.50 & 0.41 & 1.00 & 1.00 & 1.00 & 1.00 \\
50 & trxB & 6 & 10 & L R & 9 & 13 & 12 & 11 & 22 & 21 & 22 & 20 & 0.50 & 0.62 & 0.50 & 0.58 & 1.00 & 1.00 & 1.00 & 1.00 \\
50 & trxB & 7 & 11 & L R W & 10 & 14 & 18 & 17 & 22 & 21 & 21 & 20 & 0.53 & 0.58 & 0.64 & 0.70 & 1.00 & 1.00 & 1.00 & 0.94 \\
\rowcolor[RGB]{255,241,204} 50 & ttdB & 6 & 10 & R & 9 & 9 & 12 & 11 & 38 & 25 & 27 & 24 & 1.00 & 0.60 & 0.67 & 0.58 & 1.00 & 1.00 & 1.00 & 1.00 \\
\rowcolor[RGB]{255,241,204} 50 & ttdB & 8 & 13 & R & 12 & 12 & 15 & 14 & 36 & 25 & 26 & 24 & 0.86 & 0.60 & 0.68 & 0.67 & 1.00 & 1.00 & 1.00 & 1.00 \\
\rowcolor[RGB]{255,241,204} 50 & ttdB & 9 & 19 & L R W & 19 & 39 & 51 & 38 & 30 & 24 & 22 & 24 & 0.83 & 0.75 & 0.73 & 0.78 & 1.00 & 1.00 & 1.00 & 1.00 \\
50 & gatA & 7 & 20 & R & 19 & 19 & 29 & 19 & 29 & 12 & 8 & 12 & 0.59 & 0.68 & 0.76 & 0.73 & 1.00 & 1.00 & 0.97 & 1.00 \\
\cmidrule(lr){1-21}
\multicolumn{5}{r}{\emph{macro}} & -- & -- & -- & -- & 28.9 & 19.7 & 19.4 & 19.2 & \textbf{0.711} & 0.642 & 0.701 & 0.662 & 1.000 & 1.000 & 0.997 & 0.994 \\
\multicolumn{5}{r}{\emph{micro}} & -- & -- & -- & -- & -- & -- & -- & -- & 0.689 & 0.665 & \textbf{0.723} & 0.685 & 1.000 & 1.000 & 0.996 & 0.994 \\
\midrule
100 & trxB & 6 & 8 & R & 7 & 7 & 10 & 7 & 91 & 35 & 39 & 35 & 0.70 & 0.39 & 0.43 & 0.32 & 1.00 & 1.00 & 1.00 & 1.00 \\
100 & ttdB & 5 & 7 & R & 6 & 8 & 8 & 6 & 94 & 34 & 40 & 35 & 1.00 & 0.50 & 0.50 & 0.29 & 1.00 & 1.00 & 1.00 & 1.00 \\
100 & ttdB & 6 & 12 & R & 11 & 13 & 13 & 11 & 86 & 34 & 38 & 35 & 0.69 & 0.45 & 0.33 & 0.28 & 1.00 & 1.00 & 1.00 & 1.00 \\
\cmidrule(lr){1-21}
\multicolumn{5}{r}{\emph{macro}} & -- & -- & -- & -- & 90.3 & 34.3 & 39.0 & 35.0 & \textbf{0.796} & 0.446 & 0.420 & 0.293 & 1.000 & 1.000 & 1.000 & 1.000 \\
\multicolumn{5}{r}{\emph{micro}} & -- & -- & -- & -- & -- & -- & -- & -- & \textbf{0.750} & 0.444 & 0.392 & 0.289 & 1.000 & 1.000 & 1.000 & 1.000 \\
\midrule
200 & uvrB & 5 & 13 & R & 12 & 15 & 15 & 14 & 188 & 64 & 80 & 63 & 0.60 & 0.34 & 0.30 & 0.30 & 1.00 & 1.00 & 1.00 & 1.00 \\
200 & torD & 5 & 14 & R & 13 & 13 & 18 & 13 & 187 & 57 & 69 & 57 & 0.52 & 0.25 & 0.31 & 0.25 & 1.00 & 1.00 & 1.00 & 1.00 \\
200 & trxB & 4 & 5 & W & 4 & 4 & 4 & 6 & 196 & 66 & 79 & 80 & 1.00 & 0.44 & 0.40 & 0.60 & 1.00 & 1.00 & 1.00 & 1.00 \\
200 & trxB & 5 & 16 & R W & 15 & 15 & 20 & 19 & 184 & 63 & 73 & 67 & 0.65 & 0.28 & 0.32 & 0.33 & 1.00 & 1.00 & 1.00 & 1.00 \\
\rowcolor[RGB]{214,234,248} 200 & ttdB & 2 & 3 & R & 2 & 2 & 2 & 2 & 198 & 197 & 152 & 140 & 1.00 & 1.00 & 1.00 & 0.67 & 1.00 & 1.00 & 1.00 & 1.00 \\
\rowcolor[RGB]{214,234,248} 200 & ttdB & 3 & 4 & R & 3 & 3 & 4 & 3 & 197 & 139 & 152 & 139 & 1.00 & 0.75 & 1.00 & 0.60 & 1.00 & 1.00 & 1.00 & 1.00 \\
\rowcolor[RGB]{214,234,248} 200 & ttdB & 4 & 8 & R & 7 & 10 & 8 & 7 & 193 & 58 & 114 & 63 & 0.64 & 0.45 & 0.47 & 0.41 & 1.00 & 1.00 & 1.00 & 1.00 \\
\rowcolor[RGB]{214,234,248} 200 & ttdB & 5 & 18 & R & 17 & 21 & 24 & 17 & 183 & 57 & 109 & 59 & 0.63 & 0.33 & 0.38 & 0.23 & 1.00 & 1.00 & 1.00 & 1.00 \\
\rowcolor[RGB]{223,242,223} 200 & gatA & 2 & 3 & R & 2 & 2 & 2 & 2 & 198 & 143 & 146 & 142 & 1.00 & 0.67 & 0.67 & 0.67 & 1.00 & 1.00 & 1.00 & 1.00 \\
\rowcolor[RGB]{223,242,223} 200 & gatA & 4 & 6 & R & 5 & 5 & 6 & 5 & 195 & 56 & 108 & 74 & 0.83 & 0.33 & 0.46 & 0.45 & 1.00 & 1.00 & 1.00 & 1.00 \\
\rowcolor[RGB]{223,242,223} 200 & gatA & 5 & 13 & R & 12 & 12 & 18 & 12 & 188 & 56 & 102 & 57 & 0.75 & 0.26 & 0.35 & 0.26 & 1.00 & 1.00 & 1.00 & 1.00 \\
\cmidrule(lr){1-21}
\multicolumn{5}{r}{\emph{macro}} & -- & -- & -- & -- & 191.5 & 86.9 & 107.6 & 85.5 & \textbf{0.784} & 0.465 & 0.515 & 0.433 & 1.000 & 1.000 & 1.000 & 1.000 \\
\multicolumn{5}{r}{\emph{micro}} & -- & -- & -- & -- & -- & -- & -- & -- & \textbf{0.662} & 0.326 & 0.362 & 0.307 & 1.000 & 1.000 & 1.000 & 1.000 \\
\bottomrule
\end{tabular}}
\end{table}

\begin{table}[H]
\setlength{\abovecaptionskip}{5pt}
\setlength{\belowcaptionskip}{8pt}
\centering
\caption{Size 50, replicate 3. One replicate's depth ladder, so $k$ and $g_*$ move only with depth. $g_*$ is the largest cluster's gene count and \emph{singletons} the share of the $k$ clusters holding one gene.}
\label{tab:ladder-50-rep3}
{\scriptsize
\setlength{\tabcolsep}{6pt}
\begin{tabular}{r r rrrr rrrr rrrr}
\toprule
& & \multicolumn{4}{c}{$g_*$} & \multicolumn{4}{c}{singletons} & \multicolumn{4}{c}{precision} \\
\cmidrule(lr){3-6}\cmidrule(lr){7-10}\cmidrule(lr){11-14}
depth & $k$ & G & L & R & W & G & L & R & W & G & L & R & W \\
\midrule
\rowcolor[RGB]{255,241,204} 6 & 10 & 38 & 25 & 27 & 24 & 0.70 & 0.30 & 0.50 & 0.20 & 1.00 & 0.60 & 0.67 & 0.58 \\
\rowcolor[RGB]{255,241,204} 8 & 13 & 36 & 25 & 26 & 24 & 0.85 & 0.46 & 0.54 & 0.38 & 0.86 & 0.60 & 0.68 & 0.67 \\
\rowcolor[RGB]{255,241,204} 9 & 19 & 30 & 24 & 22 & 24 & 0.89 & 0.74 & 0.63 & 0.68 & 0.83 & 0.75 & 0.73 & 0.78 \\
\bottomrule
\end{tabular}}
\end{table}

\begin{table}[H]
\setlength{\abovecaptionskip}{5pt}
\setlength{\belowcaptionskip}{8pt}
\centering
\caption{Size 200, replicate 3. The same ladder at the size where the methods separate. $g_*$ is the largest cluster's gene count and \emph{singletons} the share of the $k$ clusters holding one gene.}
\label{tab:ladder-200-rep3}
{\scriptsize
\setlength{\tabcolsep}{6pt}
\begin{tabular}{r r rrrr rrrr rrrr}
\toprule
& & \multicolumn{4}{c}{$g_*$} & \multicolumn{4}{c}{singletons} & \multicolumn{4}{c}{precision} \\
\cmidrule(lr){3-6}\cmidrule(lr){7-10}\cmidrule(lr){11-14}
depth & $k$ & G & L & R & W & G & L & R & W & G & L & R & W \\
\midrule
\rowcolor[RGB]{214,234,248} 2 & 3 & 198 & 197 & 152 & 140 & 0.67 & 0.33 & 0.33 & 0.33 & 1.00 & 1.00 & 1.00 & 0.67 \\
\rowcolor[RGB]{214,234,248} 3 & 4 & 197 & 139 & 152 & 139 & 0.75 & 0.25 & 0.25 & 0.50 & 1.00 & 0.75 & 1.00 & 0.60 \\
\rowcolor[RGB]{214,234,248} 4 & 8 & 193 & 58 & 114 & 63 & 0.88 & 0.12 & 0.12 & 0.38 & 0.64 & 0.45 & 0.47 & 0.41 \\
\rowcolor[RGB]{214,234,248} 5 & 18 & 183 & 57 & 109 & 59 & 0.94 & 0.33 & 0.50 & 0.33 & 0.63 & 0.33 & 0.38 & 0.23 \\
\bottomrule
\end{tabular}}
\end{table}

\begin{table}[H]
\setlength{\abovecaptionskip}{5pt}
\setlength{\belowcaptionskip}{8pt}
\centering
\caption{Size 200, replicate 4. $g_*$ is the largest cluster's gene count and \emph{singletons} the share of the $k$ clusters holding one gene.}
\label{tab:ladder-200-rep4}
{\scriptsize
\setlength{\tabcolsep}{6pt}
\begin{tabular}{r r rrrr rrrr rrrr}
\toprule
& & \multicolumn{4}{c}{$g_*$} & \multicolumn{4}{c}{singletons} & \multicolumn{4}{c}{precision} \\
\cmidrule(lr){3-6}\cmidrule(lr){7-10}\cmidrule(lr){11-14}
depth & $k$ & G & L & R & W & G & L & R & W & G & L & R & W \\
\midrule
\rowcolor[RGB]{223,242,223} 2 & 3 & 198 & 143 & 146 & 142 & 0.67 & 0.33 & 0.33 & 0.33 & 1.00 & 0.67 & 0.67 & 0.67 \\
\rowcolor[RGB]{223,242,223} 4 & 6 & 195 & 56 & 108 & 74 & 0.83 & 0.17 & 0.17 & 0.33 & 0.83 & 0.33 & 0.46 & 0.45 \\
\rowcolor[RGB]{223,242,223} 5 & 13 & 188 & 56 & 102 & 57 & 0.92 & 0.31 & 0.38 & 0.31 & 0.75 & 0.26 & 0.35 & 0.26 \\
\bottomrule
\end{tabular}}
\end{table}

\end{document}